\documentclass[lettersize,journal]{IEEEtran}
\usepackage{amsmath,amsfonts}
\usepackage{algorithmic}
\usepackage{array}
\usepackage{textcomp}
\usepackage{stfloats}
\usepackage{url}
\usepackage{verbatim}
\usepackage{graphicx}
\def\BibTeX{{\rm B\kern-.05em{\sc i\kern-.025em b}\kern-.08em
    T\kern-.1667em\lower.7ex\hbox{E}\kern-.125emX}}
\usepackage{balance}
\usepackage{xcolor}
\usepackage{times}
\usepackage{epsfig}
\usepackage{amssymb}
\usepackage[accsupp]{axessibility}  
\usepackage{pifont}
\usepackage{wrapfig}
\usepackage{diagbox}
\usepackage{bm}
\usepackage{adjustbox}
\usepackage{booktabs}
\usepackage{cases}
\usepackage{multirow}
\usepackage{enumitem}
\usepackage[thicklines]{cancel}
\PassOptionsToPackage{prologue,dvipsnames}{xcolor}
\usepackage{textcomp,booktabs}
\usepackage{colortbl}
\usepackage{pifont}
\usepackage[numbers]{natbib}
\usepackage{subcaption}
\usepackage{titletoc}
\usepackage{tocloft}

\definecolor{mygray}{gray}{.95}
\definecolor{CColor}{rgb}{0.01,0.31,0.59}
\definecolor{GGray}{rgb}{0.80,0.90,1}
\definecolor{Shady}{rgb}{0.9,0.9,0.9}
\definecolor{kaistblue}{RGB}{20,135,200}
\definecolor{kaistdarkblue}{RGB}{0,65,145}
\definecolor{urbanablue}{RGB}{19,41,75}
\definecolor{urbanaorange}{RGB}{232,74,39}
\definecolor{drp}{rgb}{0.53,0.15,0.34}
\definecolor{darkgrey}{rgb}{0.53,0.53,0.53}
\definecolor{mygrey}{rgb}{0.9,0.9,0.9}
\definecolor{purple}{RGB}{230, 227, 254}
\definecolor{lightgreen}{RGB}{238, 252, 241}
\definecolor{lightred}{RGB}{231, 187, 187}
\definecolor{darkred}{RGB}{198, 129, 129}
\definecolor{tabhighlight}{HTML}{e5e5e5}
\definecolor{someorange}{rgb}{0.773,0.353,0.067}
\definecolor{someblue}{rgb}{0.27, 0.35, 0.760}
\usepackage[breaklinks=true,letterpaper=true,colorlinks,bookmarks=false]{hyperref}

\usepackage{amsmath}
\usepackage{amssymb}
\usepackage{mathtools}
\usepackage{amsthm}
\usepackage{bm}
\usepackage{sidecap}
\usepackage[most]{tcolorbox}  
\usepackage[capitalize,noabbrev]{cleveref}

\theoremstyle{plain}
\newtheorem{theorem}{Theorem}[section]

\theoremstyle{definition}
\newtheorem{definition}[theorem]{Definition}

\theoremstyle{remark}

\newcommand{\update}[1]{{\color{magenta}{\sf [Wait for updates!]}}}

\newcommand{\cmethodname}{\texttt{Surgery}}
\newcommand{\methodname}{\texttt{SurgeryV2}}

\begin{document}
\title{SurgeryV2: Bridging the Gap Between Model Merging and Multi-Task Learning with Deep Representation Surgery}
\author{
Enneng Yang,
Li Shen, 
Zhenyi Wang, 
Guibing Guo, 
Xingwei Wang,
Xiaochun Cao, \\
Jie Zhang,
and Dacheng Tao,~\IEEEmembership{Fellow,~IEEE}
\thanks{
This paper is an extended version of our previous work~\cite{Surgery_ICML2024} presented at ICML 2024. In this extended version, we re-analyze the limitations of the original approach and propose a new scheme, {\methodname}, which delivers improved performance, reduced data requirements, and faster training speed compared to the previous method.}

\thanks{
Enneng Yang, Guibing Guo, and Xingwei Wang are with Northeastern University, China. E-mail: ennengyang@stumail.neu.edu.cn, \{guogb,wangxw\}@swc.neu.edu.cn.}
\thanks{
Li Shen and Xiaochun Cao are with School of Cyber Science and Technology, Shenzhen Campus of Sun Yat-sen University, Shenzhen 518107, China. E-mail: mathshenli@gmail.com, caoxiaochun@mail.sysu.edu.cn.}
\thanks{
Zhenyi Wang is with the University of Maryland, USA. E-mail: wangzhenyineu@gmail.com.}
\thanks{
Jie Zhang and Dacheng Tao are with Nanyang Technological University, Singapore. E-mail: zhangj@ntu.edu.sg, dacheng.tao@gmail.com.
}}


\maketitle

\begin{abstract}
Model merging-based multitask learning (MTL) offers a promising approach for performing MTL by merging multiple expert models without requiring access to raw training data. However, in this paper, we examine the merged model's representation distribution and uncover a critical issue of ``representation bias". This bias arises from a significant distribution gap between the representations of the merged and expert models, leading to the suboptimal performance of the merged MTL model.
To address this challenge, we first propose a representation
surgery solution called {\cmethodname}.
{\cmethodname} is a lightweight, task-specific module that aligns the final layer representations of the merged model with those of the expert models, effectively alleviating bias and improving the merged model's performance.
Despite these improvements, a performance gap remains compared to the traditional MTL method.
Further analysis reveals that representation bias phenomena exist at each layer of the merged model, and aligning representations only in the last layer is insufficient for fully reducing systemic bias because biases introduced at each layer can accumulate and interact in complex ways. To tackle this, we then propose a more comprehensive solution, deep representation surgery (also called {\methodname}), which mitigates representation bias across all layers, and thus bridges the performance gap between model merging-based MTL and traditional MTL.
Finally, we design an unsupervised optimization objective to optimize both the {\cmethodname} and {\methodname} modules. Our experimental results show that incorporating these modules into state-of-the-art (SOTA) model merging schemes leads to significant performance gains. Notably, our {\methodname} scheme reaches almost the same level as individual expert models or the traditional MTL model. 
The code is available at \url{https://github.com/EnnengYang/SurgeryV2}.
\end{abstract}


\section{Introduction}
\label{sec:introduction}

\begin{figure*}[h]
\vspace{-10pt}
    \centering 
     \includegraphics[width=.75\textwidth]{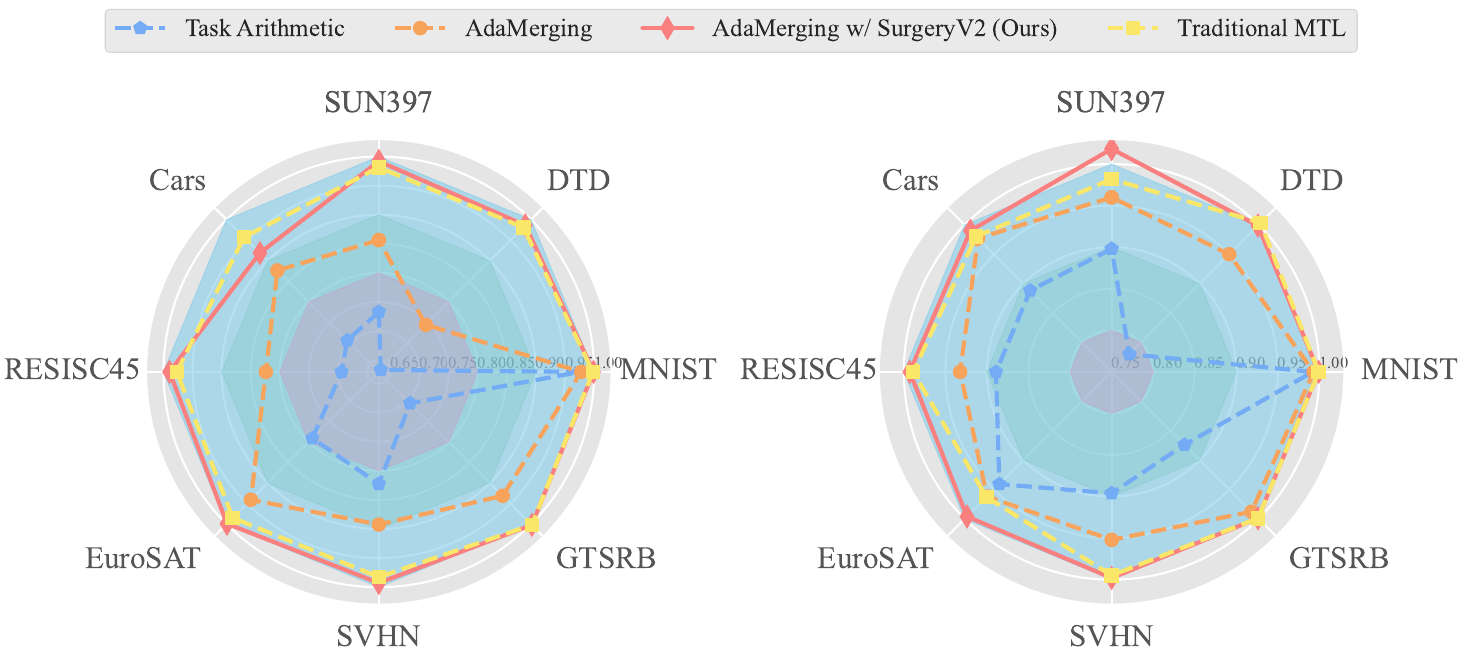}
    \vspace{-7pt}
    \caption{The {\methodname} scheme achieves performance comparable to traditional MTL when merging eight benchmark tasks and two architectures (left: ViT-B/32, right: ViT-L/14). We report the performance normalized with the individual expert model’s performance, which is denoted by the white circle of radius $1$.}   
\label{fig:sota_example} 
\vspace{-15pt}
\end{figure*}

Multi-task learning (MTL) leverages a single model to predict multiple related tasks, thereby reducing parameter costs and facilitating cross-task knowledge transfer~\cite{mtlsurvey_tpami2021,mtl_survey_acm2021,transferlearning_survey_tkde2009,mtlsurvery_recsys2018}. MTL has been extensively applied across various domains, including computer vision~\cite{gradnorm_icml2018,graddrop_neurips2020,CAGrad_NeurIPS2021,adatask_aaai2023,dwa_cvpr2019,mtlasmooSenerK18_neurips2018}, natural language processing~\cite{mtl_nlp_icml2008,mtl_nlp_translation_acl2015}, recommendation systems~\cite{mmoe_kdd2018,ple_recsys2020,MetaBalanceWWW2022} and speech recognition~\cite{mtl_speechrecognition2020,cai2021speech_2021}. However, the traditional MTL paradigm requires pre-collection of training data for all tasks and then collaboratively training an MTL model, which may lead to privacy leakage of valuable data and additional data management costs, particularly when dealing with large datasets or numerous tasks.

Recently, a novel approach known as “model merging” or “model fusion”~\cite{FisherMerging_NeurIPS2022,TangentSpace_NeurIPS2023,TiesMerging_NeurIPS2023,ZipIt_ICLR2023,pem_neurIPS2023,RegMean_ICLR2023,rewardedsoups_neurips2023,AdaMerging_ICLR2024,trainingfree_cvpr2024,LinearizationLoRA_ICLR2024,DARE_Arxiv2023,localizing_ICML2024,huang2024emr,modelfusion_survey2023,Survery_ModelMerging_2024} has emerged within the machine learning community, demonstrating promise in addressing these challenges.
The goal of model merging is to directly merge multiple expert models that have been trained individually on various tasks, to perform MTL without accessing their raw training data. In other words, model merging relies solely on the trained model parameters for each task, eliminating the need for centralized management of MTL training data and joint training. This innovative approach significantly broadens the potential applications of MTL.
Despite these advancements ($\S$\ref{subsec:relatedwork_modelmerging}), a notable performance gap persists between the most advanced model merging-based MTL methods~\cite{izmailov2018averaging,Modelsoups_ICML2022,RegMean_ICLR2023,TiesMerging_NeurIPS2023,TangentSpace_NeurIPS2023,TiesMerging_NeurIPS2023,AdaMerging_ICLR2024} and traditional MTL or individual expert models. This gap motivates our investigation into the underlying issues of model merging and develops solutions to close this gap.

In this paper, we revisit several representative model merging schemes~\cite{izmailov2018averaging,TaskArithmetic_ICLR2023,TiesMerging_NeurIPS2023,AdaMerging_ICLR2024} from the perspective of the representation distribution, identifying a common challenge: ``representation bias" ($\S$\ref{subsec:rethinking_bias}). Representation bias is defined as the gap between the representations extracted by the merged model and those extracted by the individual models. Recall that the primary goal of model merging is to create a unified model that retains the capabilities of all individual expert models. Therefore, to quantify the alignment between the merged model's representations and those of the expert models, we introduce a novel metric and conduct extensive evaluations across eight tasks (e.g., SUN397~\citep{xiao2016sun}, Cars~\citep{krause20133d}, etc.), three architectures (i.e., ViT-B/32, ViT-B/16, and ViT-L/14~\cite{Vit_ICLR2021}), and four representative model merging methods (i.e., Weight Averaging~\cite{izmailov2018averaging}, Task Arithmetic~\cite{TaskArithmetic_ICLR2023}, Ties-Merging~\cite{TiesMerging_NeurIPS2023}, AdaMerging~\cite{AdaMerging_ICLR2024}).
By examining the relationship between the representation bias of the merged model and its final MTL performance, we empirically demonstrate that representation bias is a critical factor limiting the performance of the merged model. Consequently, reducing representation bias can achieve better model merging performance.

To solve the ``representation bias” problem in model merging, we first propose a novel representation surgery solution ($\S$\ref{subsec:surgery}), termed {\cmethodname}, which aligns the representation of the merged model at the \textit{last layer} with the representation of the individual expert model at the \textit{last layer}. Since raw training data are not available, we design an unsupervised surrogate objective to update parameters in the {\cmethodname} module.
Notably, this approach achieves a new SOTA performance among existing model merging methods. However, there is still a relatively obvious representation bias after using the representation surgery scheme, and the MTL performance of the merged model is still far behind the traditional MTL method. This result prompted us to conduct a deeper analysis of the representation surgery method to further reduce representation bias and narrow the performance gap.

We carefully analyze the correlation between the capacity of the {\cmethodname} module and the degree of representation bias ($\S$\ref{subsec:revisiting}), finding that simply increasing the capacity can effectively reduce the bias and improve merging performance. However, beyond a certain threshold, further increasing capacity no longer yields benefits. Additionally, our layer-by-layer analysis reveals that representation bias exists at each layer of the merged model, with the bias magnitude increasing monotonically with the depth of the layer in the merged model. 
In summary, representation bias exists at every layer of the merged model, and addressing only the last layer—even with a substantial capacity—fails to fully resolve systemic representation bias. This is because the biases introduced at each layer can accumulate and interact in complex ways.
To overcome this limitation, we then propose a deep representation surgery solution  ($\S$\ref{subsec:ourmethod}), abbreviated as {\methodname}, which effectively alleviates layer-wise representation bias and bridges the performance gap between model
merging based MTL and traditional MTL. Specifically, our {\methodname} aligns the representation of the merged model with that of the individual models at \textit{each layer}, thereby effectively mitigating MTL performance degradation.

We conduct numerous experiments to demonstrate the {SOTA performance} of our {\cmethodname} and {\methodname} schemes ($\S$\ref{sec:experiment}).  Our comprehensive evaluation covered eight vision datasets, five text datasets, four representative model merging methods, and four model architectures within the computer vision (CV) and natural language processing (NLP) domains.
Our observations include the following: (i) when {\cmethodname} and {\methodname} schemes are applied to existing model merging methods, the merged model achieves significantly better MTL performance. In particular, as shown in Fig.~\ref{fig:sota_example}, our {\methodname} model merging scheme achieves comparable or even slightly better performance than the traditional MTL method, which is noteworthy. 
(ii) By monitoring the performance changes during the optimization process, we find that with the same capacity, {\methodname} can achieve significantly higher accuracy than {\cmethodname} with fewer iterations. 
(iii) Our proposed {\methodname} is effective even in wild data settings, significantly broadening its range of application scenarios.

The \textbf{main contributions} of this paper are as follows:
\begin{itemize}
[noitemsep,topsep=0pt,parsep=0pt,partopsep=0pt,leftmargin=*]
     \item We revisit representative model merging methods and, for the first time, identify the ``representation bias" problem (which exists across tasks, architectures, and merging methods) as a major cause of poor MTL performance. 
     \item We propose a novel ``representation surgery" approach, termed {\cmethodname}, to mitigate ``representation bias" in the merged model. Notably, the {\cmethodname} module is suitable for any existing model merging algorithm.
     \item We reveal the correlation between the capacity of {\cmethodname} module and the degree to which representation bias is resolved. Furthermore, we find that representation bias occurs in each layer of the merged model, and addressing this bias solely in the final layer is insufficient.
     \item We introduce a deep representation surgery scheme (also called {\methodname}), which applies representation surgery to each layer of the merged model, effectively reducing layer-wise representation bias.
     \item We conduct extensive experiments across CV and NLP domains, including eight vision datasets, five text datasets, four model architectures, and four representative model merging methods. The results substantiate the effectiveness of both our {\cmethodname} and {\methodname}. 
\end{itemize}

\section{Related Works}
\label{sec:relatedworks}
\label{subsec:relatedwork_modelmerging}

In this section, we briefly introduce the related work, with a more comprehensive discussion provided in Appendix~\ref{sec:relatedworks_appendix}.

Recent studies have attempted to achieve MTL using model merging techniques~\cite{modelfusion_survey2023,ZipIt_ICLR2023,akiba2024evolutionary,merging_gradientmarching_iclr2024,Survery_ModelMerging_2024}. However, the performance of simple parameter averaging~\cite{izmailov2018averaging,mcmahan2017communication} degrades significantly as the number of tasks increases, resulting in a substantial gap between its performance and that of traditional MTL model.
Many recent works have made various attempts to fill this gap~\cite{TaskArithmetic_ICLR2023,TiesMerging_NeurIPS2023,DARE_Arxiv2023,RegMean_ICLR2023,wu2023pi,rewardedsoups_neurips2023,pem_neurIPS2023,daheim2024model,AdaMerging_ICLR2024,Survery_ModelMerging_2024}. 
The \textit{first type} of method explores how to better weigh and combine multiple models~\cite{FisherMerging_NeurIPS2022,zhou2024metagpt,akiba2024evolutionary}. For example, Fisher-Merging~\cite{FisherMerging_NeurIPS2022} performs weighted merging utilizing the importance of each parameter through the Fisher information matrix~\cite{fisher1922mathematical}. RegMean~\cite{RegMean_ICLR2023} reweights and linearly combines rows in weight matrices based on statistics from training data.
AdaMerging~\cite{AdaMerging_ICLR2024} leverages unlabeled test data to automatically learn a set of task-level or layer-level model merging coefficients.
The \textit{second type} of method explores how to merge models in sparse subspaces to reduce interference~\cite{panigrahi2023task,du2024NeurIPSparameter,localizing_ICML2024,zimmer2024ICLRsparse,deep2024della,huang2024emr,davari2023model}. For example, Ties-Merging~\cite{TiesMerging_NeurIPS2023} removes the smaller magnitude parameters and eliminates the issue of parameter sign conflicts during model merging. DARE~\cite{DARE_Arxiv2023} removes a large number of useless neuron updates and then scales the neurons for merging. 
Concrete~\cite{tang2023concrete} finds a shared subspace between multiple tasks for model merging. 
The \textit{third type} of method dynamically merges multiple expert modules during inference~\cite{tang2024merging,muqeeth2024learning,li2024ICLRmerge,lu2024twin,yadav2024survey}. For example, WEMoE~\cite{tang2024merging} dynamically merges linear layers by routing, and static merges nonlinear layers. It should be noted that the dynamic merging method requires additional maintenance of more model parameters than the first two categories of methods, and it also reduces the inference efficiency.

While existing methods predominantly concentrate on merging in weight space, they often neglect a crucial concern stemming from weight merging--the \textit{representation bias}. A substantial disparity emerges in the \textit{representation space} between the merged model and individually trained expert models. In contrast, our method addresses this gap, aiming to minimize the representation discrepancy. Moreover, our approach operates in the \textit{representation space}, offering a complementary and orthogonal perspective to traditional weight-space merging methods. Consequently, our method can be seamlessly integrated with them.

\section{Preliminary}

In this section, we first define the notation within the context of model merging ($\S$\ref{subsec:setting}), and then introduce some representative model merging solutions ($\S$\ref{subsec:preliminary}).

\subsection{Notations and Problem Setting}
\label{subsec:setting}

\textbf{Notations}: Assume there are $T$ models ($\small \Phi_{\mathbf{\Theta}^{(1)}},\ldots,\Phi_{\mathbf{\Theta}^{(T)}}$) of the same architecture that needs to be merged, and they are fine-tuned into a pretrained model $\small \Phi_{\mathbf{\Theta}^{(0)}}$ on tasks $1-T$ respectively. The parameters (or weights) of the $t$-th model $\small \Phi_{\mathbf{\Theta}^{(t)}}$ are represented as $\small \mathbf{\Theta}^{(t)}=\{\mathbf{\Theta}^{(t)}_l\}_{l=1}^L$, where $l$ denotes the $l$-th layer~\footnote{\footnotesize This paper uses the terms of layer, block and module interchangeably, and they all refer to same concept.} of the model, and $L$ is the total number of layers. In addition, the test set of the $t$-th task is defined as $\small \mathcal{D}_{te}^{(t)}=\{\mathcal{X}^{(t)},\mathcal{Y}^{(t)}\}=\{x_i^{(t)}, y_i^{(t)}\}_{i=1}^{|\mathcal{D}_{te}^{(t)}|}$, where $x_i^{(t)}$ represents the $i$-th input sample and $y_i^{(t)}$ represents the corresponding label, and $|\mathcal{D}_{te}^{(t)}|$ is the number of samples. Without loss of generality, the forward propagation of a deep neural network model can be represented as a directed acyclic computation graph, where the output of each layer serves as the input for the subsequent layer. 
Based on these notations, we give definitions of representations and model merging in Definition \ref{def:representation} and Definition \ref{def:model_merging}, respectively.

\begin{definition}[\textbf{Representation}]
\label{def:representation}
Given a model $\small \Phi_{\mathbf{\Theta}^{(t)}}$, we define the output (also referred to as \textit{representation}) of sample $x_i^{(t)}$ at the $l$-th layer as $z^{(t)}_{l,i}=\Phi_{\mathbf{\Theta}^{(t)}_{1:l}}(x_i^{(t)}) \in \mathbb{R}^{d_l}$, and $d_l$ represents the output dimension of the $l$-th layer. Similarly, the representation of all samples for task $t$ at the $l$-th layer is denoted by $\mathbf{Z}^{(t)}_l=\{z^{(t)}_{l,i}\}_{i=1}^{|\mathcal{D}_{te}^{(t)}|} \in \mathbb{R}^{d_l \times |\mathcal{D}_{te}^{(t)}|}$. 
\end{definition}

\begin{definition}[\textbf{Model Merging}]
\label{def:model_merging}
Given multiple expert models fine-tuned on different tasks, the goal of model merging is to merge the parameters $\small \{\mathbf{\Theta}^{(1)},\ldots,\mathbf{\Theta}^{(T)}\}$, and finally obtain the new parameters $\small \mathbf{\Theta}^{(mtl)}=\texttt{merge}(\mathbf{\Theta}^{(1)},\ldots,\mathbf{\Theta}^{(T)})$, and $\Phi_{\mathbf{\Theta}^{(mtl)}}$ can successfully perform tasks $1-T$. 
In other words, the objective of model merging is for the MTL model $\small \Phi_{\mathbf{\Theta}^{(mtl)}}$ to achieve the lowest loss on test sets of all tasks, which is defined as follows:
\begin{equation}
\small 
\begin{split}
    \arg \min_{\mathbf{\Theta}^{(mtl)}} \; \frac{1}{T}\sum_{t=1}^T \sum_{i=1}^{|\mathcal{D}_{te}^{(t)}|} \frac{1}{|\mathcal{D}_{te}^{(t)}|} \mathcal{L}^{(t)} \left(\Phi_{\mathbf{\Theta}^{(mtl)}}\left(x_i^{(t)}\right), y_i^{(t)}\right), 
    \; \\  \text{where }\mathbf{\Theta}^{(mtl)}=\texttt{merge}\left(\mathbf{\Theta}^{(1)},\ldots,\mathbf{\Theta}^{(T)}\right),
\end{split}
\label{eq:merging_obj}
\end{equation}
where $\mathcal{L}^{(t)}(\cdot)$ is the loss function for the $t$-th task, such as cross-entropy loss or mean squared error. 
\end{definition}

Taking two tasks as an example, as illustrated in Fig.~\ref{fig:method}, model merging hopes to combine the expert models trained individually on task $A$ and task $B$ (i.e., Fig.~\ref{fig:method}(a)) to obtain a new model (i.e., Fig.~\ref{fig:method}(b)) capable of performing both tasks $A$ and $B$. Mainstream model merging research focuses on designing more advanced $\texttt{merge}(\cdot)$ strategies to alleviate task conflict and task interference~\cite{Modelsoups_ICML2022,TaskArithmetic_ICLR2023,RegMean_ICLR2023,TiesMerging_NeurIPS2023,DARE_Arxiv2023,AdaMerging_ICLR2024,du2024NeurIPSparameter}.

\subsection{Representative Model Merging Methods}
\label{subsec:preliminary}

In this section, based on the popularity and performance of model merging methods, we select four representative methods to introduce: Weighted Averaging~\cite{izmailov2018averaging}, Task Arithmetic~\cite{TaskArithmetic_ICLR2023}, Ties-Merging~\cite{TiesMerging_NeurIPS2023}, and AdaMerging~\cite{AdaMerging_ICLR2024}.

When merging models, one of the simplest solutions is \textbf{\textit{Weighted Averaging}}, defined as $\small \mathbf{\Theta}^{(mtl)} = \frac{1}{T} \sum_{t=1}^T \mathbf{\Theta}^{(t)}$. However, it often results in suboptimal performance due to potential task conflicts.
As mentioned in $\S$\ref{subsec:relatedwork_modelmerging}, a lot of advanced work has been devoted to developing better merge operations (i.e., the $\texttt{merge}(\cdot)$ function in Eq.~\ref{eq:merging_obj}) to reduce the performance degradation of the merged model $\Phi_{\mathbf{\Theta}^{(mtl)}}$ compared to individual models $\small \{\Phi_{\mathbf{\Theta}^{(1)}},\ldots,\Phi_{\mathbf{\Theta}^{(T)})\}}$ or the traditional MTL model.
As a representative work, \textbf{\textit{Task Arithmetic}}~\cite{TaskArithmetic_ICLR2023} uses a grid search algorithm to find a better model merging coefficient $\lambda$, i.e., 
$\small \mathbf{\Theta}^{(mtl)} = \mathbf{\Theta}^{(0)} + \lambda \cdot \sum_{t=1}^T (\mathbf{\Theta}^{(t)} - \mathbf{\Theta}^{(0)})$, where $\small \mathbf{\Theta}^{(0)}$ is the pre-trained model's weight.
Building on this approach, \textbf{\textit{Ties-Merging}}~\cite{TiesMerging_NeurIPS2023} suggests that eliminating parameter sign conflicts prior to merging can reduce inter-model interference. Its merge rule is $\small \mathbf{\Theta}^{(mtl)} = \mathbf{\Theta}^{(0)} + \lambda \cdot \sum_{t=1}^T \phi(\mathbf{\Theta}^{(t)} - \mathbf{\Theta}^{(0)})$, where $\phi(\cdot)$is a function that resolves sign conflicts. 
Subsequently, \textbf{\textit{AdaMerging}}~\cite{AdaMerging_ICLR2024} emphasizes the critical impact of merging coefficients on performance. It proposes optimizing the layer-level merging coefficients $\lambda_l^{(t)}$ formalized as $\small \mathbf{\Theta}^{(mtl)} = \{\mathbf{\Theta}^{(0)}_l + \sum_{t=1}^T \lambda_l^{(t)} 
 \cdot (\mathbf{\Theta}^{(t)}_l - \mathbf{\Theta}^{(0)}_l)\}_{l=1}^L$.

These advanced works have achieved more powerful performance than the vanilla Weighted Averaging method. However, there is still a significant performance gap compared with individual models or the traditional MTL model. This motivates us to further analyze the reasons behind the limited performance of these model merging methods.

\section{Method}
\label{sec:method}

This section first re-examines these representative merging schemes from the perspective of representation distribution and establishes that they suffer from ``representation bias'' ($\S$\ref{subsec:rethinking_bias}).
Then, we propose a representation surgery (i.e., {\cmethodname}) method to mitigate representation bias ($\S$\ref{subsec:surgery}). Despite the new SOTA performance achieved by {\cmethodname}, it still has gaps with traditional MTL or expert models.
Next, we conduct an in-depth analysis of the {\cmethodname} scheme and reveal the phenomenon of layer-wise representation bias ($\S$\ref{subsec:revisiting}).
Finally, we propose a deep representation surgery scheme (i.e., {\methodname}) scheme to perform representation alignment at each layer of the merged model ($\S$\ref{subsec:ourmethod}).

\begin{figure*}[t]
    \centering 
        \begin{minipage}[t]{0.32\linewidth}
         \includegraphics[width=.49\textwidth]{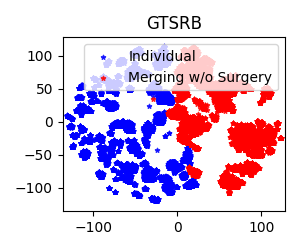
        }
        \includegraphics[width=.49\textwidth]{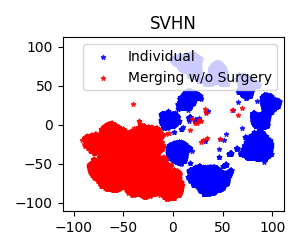
        }
        \vspace{-20pt}
        \begin{center}
            \text{\small (a) Weight Averaging on ViT-B/32}
         \end{center}
        \end{minipage}
        \begin{minipage}[t]{0.32\linewidth}
         \includegraphics[width=.49\textwidth]{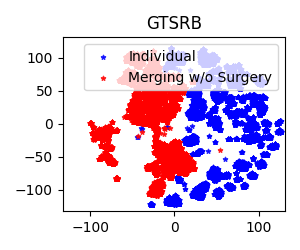}
        \includegraphics[width=.49\textwidth]{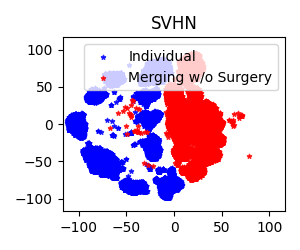}
        \vspace{-20pt}
        \begin{center}
            \text{\small (b) Task Arithmetic on ViT-B/32}
         \end{center}
        \end{minipage}
        \begin{minipage}[t]{0.32\linewidth}
          \includegraphics[width=.49\textwidth]{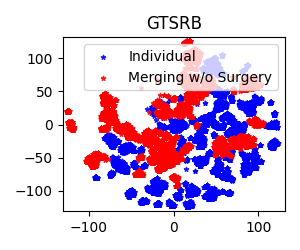}
        \includegraphics[width=.49\textwidth]{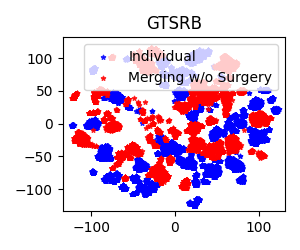}
        \vspace{-20pt}
        \begin{center}
            \text{\small (c) AdaMerging on ViT-B/32 and ViT-B/16}
         \end{center}
        \end{minipage}
    \caption{Visualization of the distribution of representations extracted by the merged model (\textcolor{red}{red}) for the existing model merging schemes and representations extracted by the individual expert model (\textcolor{blue}{blue}). We observe that there is a clear distribution discrepancy between the two.}  
\label{fig:dis_wo_surgery} 
\vspace{-15pt}
\end{figure*}

\subsection{Revisiting Representation Bias}
\label{subsec:rethinking_bias}

This section analyzes representative merging schemes mentioned in $\S$\ref{subsec:preliminary} from the perspective of the representation distribution of the merged model and expert models and identifies a key issue that limits the MTL performance of the merged model, namely, ``representation bias". We give a clear definition of representation bias in Definition \ref{def:representation_bias}.

\begin{definition}[\textbf{Representation Bias}]
\label{def:representation_bias}
Given merged model $\small \Phi_{\Theta^{(mtl)}}$ and individual expert models $\small \{\Phi_{\mathbf{\Theta}^{(1)}},\ldots,\Phi_{\mathbf{\Theta}^{(T)}}\}$, the representation bias $b^{(t)}_L$ w.r.t. task $t$ is defined as the discrepancy between the representations $\small \mathbf{Z}_{L}^{ (t), ind}$ and $\small \mathbf{Z}_{L}^{(t), mtl}$ extracted from the \textit{final} (i.e., $L$-th) layer of the individual expert model $\small \Phi_{\mathbf{\Theta}^{(t)}}$ and the merged model $\small \Phi_{\mathbf{\Theta}^{(mtl)}}$:
\begin{equation}
\small
    b^{(t)}_L = \frac{1}{|\mathcal{D}_{te}^{(t)}|}  \frac{1}{d_L} \;\; \mathcal{\psi} \left(\mathbf{Z}_{L}^{(t), mtl},\; \mathbf{Z}_{L}^{ (t), ind}\right),
\label{eq:l1distance}
\end{equation} 
where $\small \psi(\cdot)$ represents any distance-measuring function, such as $L_1$ or $L_2$ loss, and $d_L$ is the dimension of the representation at the final layer (e.g., ViT-B/32 and ViT-B/16 are 512, and ViT-L/14 is 768). 
\end{definition}

\begin{figure}[t]
\vspace{-10pt}
    \centering 
    \includegraphics[width=0.45\textwidth]{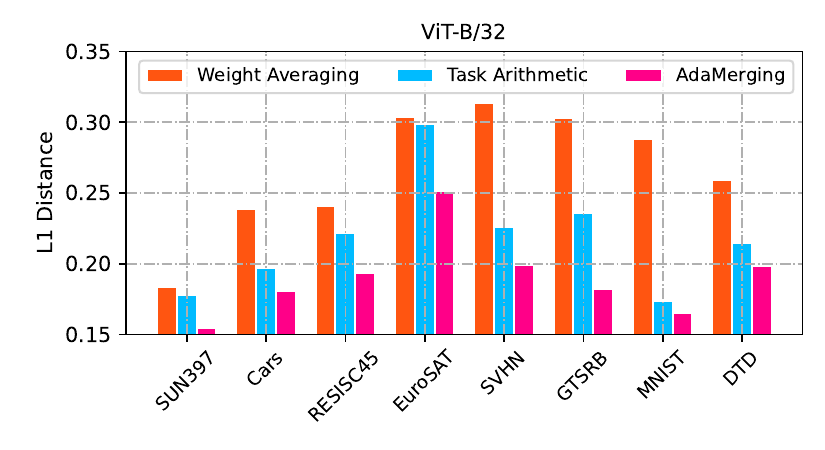}
    \vspace{-10pt}
    \caption{The ``representation bias'' (in Eq.~\ref{eq:l1distance} of Definition \ref{def:representation_bias}) between representations extracted by the merged model using three model merging methods and representations extracted by individual expert models. 
    }  
\label{fig:l1_distance_compare} 
\vspace{-10pt}
\end{figure}

Without loss of generality, we perform analysis on
the following representative model merging methods,
tasks/datasets, and architectures: 
(i) \textit{\textbf{Methods}}: We analyze the four model merging methods mentioned in $\S$\ref{subsec:preliminary}.
(ii) \textit{\textbf{Tasks}}: We follow Task Arithmetic~\cite{TaskArithmetic_ICLR2023} and use the following eight datasets as eight tasks for model merging: SUN397, Cars, RESISC45, EuroSAT, SVHN, GTSRB, MNIST, DTD. We provide a detailed dataset description in Appendix~\ref{sec:experimental_details_appendix}.
(iii) \textit{\textbf{Architectures}}: We merge eight expert models into one model and experiment with three ViT architectures~\cite{Vit_ICLR2021} with different parameter scales: ViT-B/32, ViT-B/16, and ViT-L/14.

\begin{figure}[t]
\vspace{-5pt}
    \centering 
    \includegraphics[width=.42\textwidth]{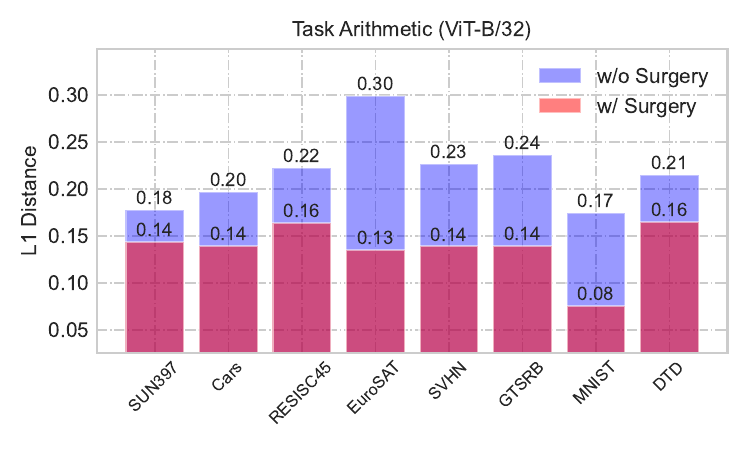}
    \vspace{-10pt}
    \caption{Visualization of the ``representation bias'' (in Eq.~\ref{eq:l1distance}) of the representation of the merged model \textbf{with} (\textcolor{red}{red}) and \textbf{without} (\textcolor{blue}{blue}) representation surgery versus the individual model.
    }  
\label{fig:l1_distance_vitb32} 
\vspace{-15pt}
\end{figure}

To reveal the performance gap between the merged model $\Phi_{\Theta^{(mtl)}}$ and the individual expert models $\small \{\Phi_{\mathbf{\Theta}^{(1)}},\ldots,\Phi_{\mathbf{\Theta}^{(T)}}\}$, we use the t-SNE~\cite{tsne2008} tool to visualize the representations $\small \mathbf{Z}_{L}^{ (t), ind}$ and $\small \mathbf{Z}_{L}^{(t), mtl}$  of the expert and merged models, as well as directly measure the representation bias in Definition \ref{def:representation_bias}. We find that ``representation bias'' exists \textbf{across tasks, across merging methods, and across architectures}:
\begin{itemize}[noitemsep,topsep=0pt,parsep=0pt,partopsep=0pt,leftmargin=*]
    \item \textit{From the perspective of the representation distribution}~\footnote{\footnotesize Due to space limitations, we only show two tasks (GTRSB and SVHN), three merging methods (Weight Averaging, Task Arithmetic, and AdaMerging), and two architectures (ViT-B/32 and ViT-B/16) in the main text.}. 
    As shown in  Fig.~\ref{fig:dis_wo_surgery}, we have the following observations: 
    (i) Across different tasks: As shown in Fig.~\ref{fig:dis_wo_surgery}(a), on the two tasks of GTSRB and SVHN, the representation distributions of the merged model (red points) and the individual expert model (blue points) are quite different.
    (ii) Across different merging methods: Comparing Fig.~\ref{fig:dis_wo_surgery} (a), (b), (c), we find that the phenomenon of inconsistent representation distribution between the merged model and the individual model exists in Weight Averaging, Task Arithmetic and AdaMerging.
    (iii) Across different architectures: In Fig.~\ref{fig:dis_wo_surgery}(c), we observe that there is also an inconsistency in the representation distribution under ViT-B/32 and ViT-B/16 architectures.
    \item \textit{From the perspective of the representation bias}. We quantify the representation bias between independent and merged models according to Definition \ref{def:representation_bias}. As shown in Fig.~\ref{fig:l1_distance_compare}, we clearly observe that AdaMerging has a smaller representation bias between the merged model and the individual models compared to the other two model merging methods, i.e., Task Arithmetic and Weight Averaging. Meanwhile, Task Arithmetic also has a smaller representation bias than the Weight Averaging method.
\end{itemize}

\begin{figure*}[t]
    \centering 
        \begin{minipage}[t]{0.32\linewidth}
         \includegraphics[width=.49\textwidth]{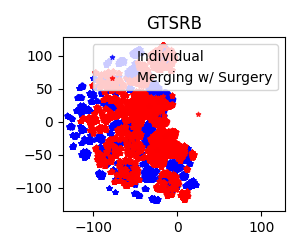
        }
        \includegraphics[width=.49\textwidth]{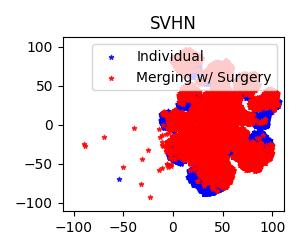
        }
        \vspace{-20pt}
        \begin{center}
            \text{\small (a) Weight Averaging on ViT-B/32}
         \end{center}
        \end{minipage}
        \begin{minipage}[t]{0.32\linewidth}
         \includegraphics[width=.49\textwidth]{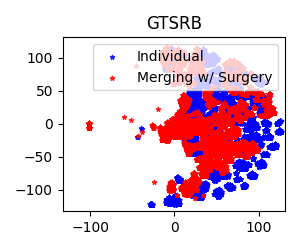}
        \includegraphics[width=.49\textwidth]{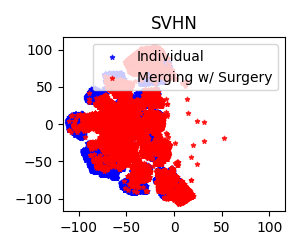}
        \vspace{-20pt}
        \begin{center}
            \text{\small (b) Task Arithmetic on ViT-B/32}
         \end{center}
        \end{minipage}
        \begin{minipage}[t]{0.32\linewidth}
         \includegraphics[width=.49\textwidth]{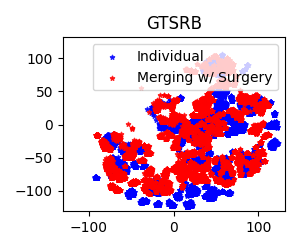}
        \includegraphics[width=.49\textwidth]{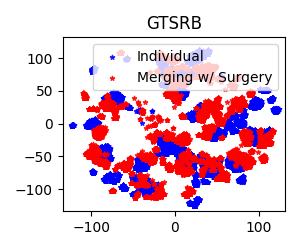}
        \vspace{-20pt}
        \begin{center}
            \text{\small (d) AdaMerging on  ViT-B/32 and ViT-B/16}
         \end{center}
        \end{minipage}
    \caption{Visualization of the distribution of features extracted by the merged model {\textbf{after performing the representation surgery}} (\textcolor{red}{red}) and features extracted by the individual model (\textcolor{blue}{blue}). We observe that the two distributions overlap highly.} 
\label{fig:dis_w_surgery} 
\vspace{-10pt}
\end{figure*}

This phenomenon inspires us to think: \textbf{Is representation bias the key factor limiting the MTL performance improvement of model merging}? We give a \textit{positive} answer.
On the one hand, previous research results~\cite{AdaMerging_ICLR2024,TiesMerging_NeurIPS2023,TaskArithmetic_ICLR2023} and the performance comparison in this paper (e.g., Tab.~\ref{tab:performance_vitbase32}, Tab.~\ref{tab:performance_vitlarge14} and Tab.~\ref{tab:performance_vitbase16_appendix} (in Appendix~\ref{subsec:performance_appendix})) show that in terms of the MTL performance of the merged model: AdaMerging $\small \!\!>\!\!$ Task Arithmetic $\small \!\!>\!\!$ Weight Averaging. On the other hand, our results in Fig.~\ref{fig:dis_wo_surgery} and Fig.~\ref{fig:l1_distance_compare} show that AdaMerging $\small \!\!<\!\!$ Task Arithmetic $\small \!\!<\!\!$ Weight Averaging, whether in terms of representation distribution or representation bias. Therefore, MTL performance and representation bias show a strong positive correlation.

The above analysis demonstrates that \textit{``representation bias" presents substantial challenges in model merging-based MTL}. The merged model's performance improves as the representation bias decreases, motivating us to develop a solution to mitigate this issue in model merging-based MTL methods.

\subsection{{\cmethodname}: Representation Surgery}
\label{subsec:surgery}

\begin{figure*}[t]
    \centering 
     \includegraphics[width=1.\textwidth]{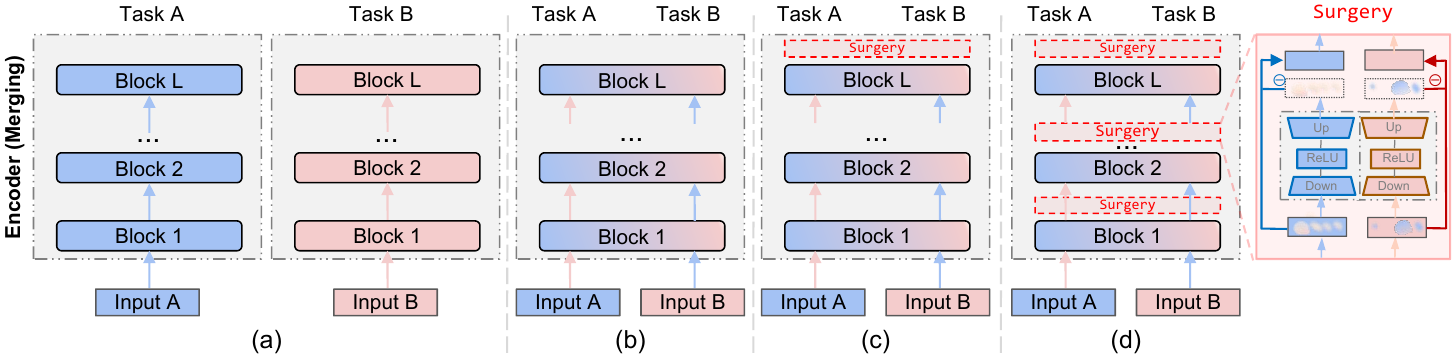}
\vspace{-10pt}
    \caption{An overview of model merging solutions: (a) individual models; (b) merged model produced by traditional model merging method (e.g., Weight Averaging, Task Arithmetic~\cite{TaskArithmetic_ICLR2023}, Ties-Merging~\cite{TiesMerging_NeurIPS2023}, AdaMerging~\cite{AdaMerging_ICLR2024}); (c) merged model employing our {\cmethodname} scheme in $\S$\ref{subsec:surgery}; (d) merged model utilizing our {\methodname} scheme in $\S$\ref{subsec:ourmethod}.
}  
\label{fig:method} 
\vspace{-10pt}
\end{figure*}

In this section, to alleviate the representation bias, a novel representation surgery scheme is proposed, which involves aligning $\mathbf{Z}_{L}^{ (t), ind}$ and $\mathbf{Z}_{L}^{(t), mtl}$ for task $t$ using a representation surgery module $\small \Omega_{\mathbf{W}^{(t)}}(\cdot)$ as Fig.~\ref{fig:method}(c). The formal objective function is defined as:
\begin{equation}
\small
\centering
   \begin{split}
     \min_{\{\mathbf{W}^{(1)}, \ldots, \mathbf{W}^{(T)}\}} \;
    \sum_{t=1}^T \frac{1}{|\mathcal{D}_{te}^{(t)}|} \mathcal{\psi}
    \left(\mathbf{\hat{Z}}_{L}^{(t), mtl}
    ,\; \mathbf{Z}_{L}^{ (t), ind}\right), \\
    \text{ where } \mathbf{\hat{Z}}_{L}^{(t), mtl} = \mathbf{Z}_{L}^{(t), mtl} - \Omega_{\mathbf{W}^{(t)}}(\mathbf{Z}_{L}^{(t), mtl}),
   \end{split}
\label{eq:surgery_v1}
\end{equation} 
where $\small \mathcal{\psi}(\cdot)$ is an arbitrary distance function, $\small \Omega_{\mathbf{W}^{(t)}}(\cdot)$ is an optimizable task-private lightweight module, which can be an arbitrary implementation (such as multiple fully connected layers, etc.). 
Without loss of generality, in this paper, we implement it as an Adapter-like structure~\cite{adapter_icml2019}, that is, 
\begin{equation}
\small
    \Omega_{\mathbf{W}^{(t)}}(\mathbf{Z}_{L}^{(t), mtl})=\mathbf{W}_{up}^{(t)} \cdot \texttt{ReLU}(\mathbf{W}_{down}^{(t)} \cdot ({\mathbf{Z}_{L}^{(t), mtl}})^{\top}),
\label{eq:surgery}
\end{equation}
where $\small \mathbf{W}_{up}^{(t)} \!\!\in\!\! \mathbb{R}^{d_L \times r}$, $\small \mathbf{W}_{down}^{(t)} \!\!\in\!\! \mathbb{R}^{r \times d_L}$ represent two learnable matrices, i.e., $\small {\mathbf{W}^{(t)}}\!=\!\{\mathbf{W}_{up}^{(t)}, \mathbf{W}_{down}^{(t)}\}$, $\texttt{ReLU}$($\cdot$) is a nonlinear activation function, $\small d_L$ is the dimension of representation as mentioned in Definition~\ref{def:representation}, and $r$ is a hyperparameter, also called rank, which we set to $16$ by default.

In particular, our {\cmethodname} scheme {does not rely on any labeled training data}. Instead, inspired by test-time adaptation~\citep{Tent_ICLR2021,StableTTA_ICLR2023}, it leverages the unlabeled test data $\small \{\mathcal{D}^t_{te}(\mathcal{X}, \mathcal{Y})\}_{t=1}^T$ and individual models $\small \{\Phi_{\Theta_t}\}_{t=1}^T$ as a self-supervised signal to optimize the parameters of the {\cmethodname} module, denoted as $\small \{\mathbf{W}^{(1)}, \mathbf{W}^{(t)}, \ldots, \mathbf{W}^{(T)}\}$.

\textbf{Discussion}. The representation surgery proposed in this paper complements existing model merging schemes. We next discuss why the proposed {\cmethodname} scheme is effective. 
As mentioned in Eq.~\ref{eq:surgery}, the goal of the representation surgery is to reduce the discrepancy in the representation distributions between the merged and individual models.
Comparing the discrepancy in representation distributions in Fig.~\ref{fig:dis_wo_surgery} (w/o {\cmethodname}) and Fig.~\ref{fig:dis_w_surgery} (w/ {\cmethodname}), we can observe that the distributions of the merged and individual models in Fig.~\ref{fig:dis_w_surgery} are closer (i.e., higher overlap). In addition, we quantify the ``representation bias'' (in Definition~\ref{def:representation_bias}) between the two distributions. As illustrated in Fig.~\ref{fig:l1_distance_vitb32}, the proposed representation surgery (red column) significantly reduces the $L_1$ distance for the merged model compared to the model merged without representation surgery (blue column). For instance, on the EuroSAT dataset, Task Arithmetic-based model merging decreases the representation bias from 0.30 to 0.13 following representation surgery, representing a 56\% relative reduction. These findings indicate that the proposed {\cmethodname} effectively mitigates the problem of representation bias.

\subsection{Revisiting Representation Surgery}
\label{subsec:revisiting}

\begin{figure}[t]
\vspace{-5pt}
    \centering 
     \includegraphics[width=0.24\textwidth]{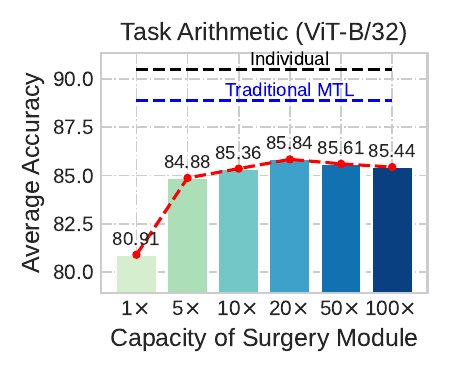}
     \includegraphics[width=0.24\textwidth]{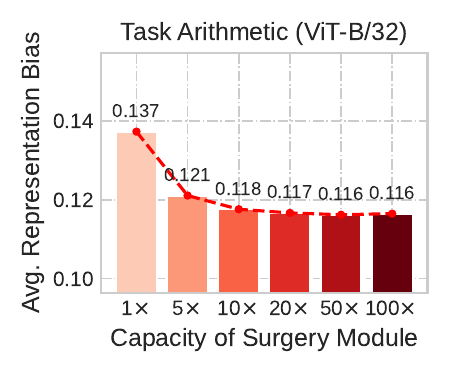}
    \vspace{-18pt}
    \caption{An illustration of changes in the merged model's {performance} (left, higher better) and {representation bias} (right, lower better) as the {\cmethodname}'s capacity increases. 
    } 
\label{fig:increase_rank_acc_bias_tv_vit_b32} 
\vspace{-15pt}
\end{figure}

The representation surgery scheme mentioned in $\S$\ref{subsec:preliminary} has indeed achieved significant performance improvements compared to existing model merging approaches. However, there is still a certain gap between it and the individual model or the traditional MTL model, irrespective of whether we consider representation distribution visualization or performance comparison. 
For example, in Fig.~\ref{fig:dis_w_surgery}, the red and blue distributions exhibit imperfect overlap. Furthermore, as depicted in the left subfigure of Fig.~\ref{fig:increase_rank_acc_bias_tv_vit_b32}, the optimal performance of Task Arithmetic with {\cmethodname} is 85.8, while the individual model is 90.5 and the traditional MTL model is 88.9. 
This encourages us to intensify our efforts towards narrowing these gaps.
In this section, by revisiting the Representation Surgery scheme (in $\S$\ref{subsec:surgery}), we discerned two key observations.

\textbf{Observation 1}: \textit{Increasing the surgery module's capacity can enhance performance and reduce representation bias, but beyond a certain threshold, additional capacity has a negligible effect.}
The {\cmethodname} module $\small \Omega_{\mathbf{W}^{(t)}}(\cdot)$ in Eq.~\ref{eq:surgery_v1} is a learnable module (e.g., a multi-layer perceptron or an Adapter-like module~\cite{adapter_icml2019}), prompting a natural conjecture that expanding its capacity could ameliorate the representation bias and enhance performance.
As depicted in Fig.~\ref{fig:increase_rank_acc_bias_tv_vit_b32}, when we increase the capacity from $1 \times$ to $20 \times$, the average performance of the merged model increases from 80.9 to 85.8 (left subplot). Concomitantly, the representation bias at the last layer also decreased from 0.13 to 0.11 (right subplot). However, further increments in capacity, such as by a factor of $100 \times$, do not yield substantial improvements in performance or bias. This suggests a ceiling (significantly lower than traditional MTL) on the efficacy of performing representation surgery at the \textit{final layer} in alleviating representation bias.

\begin{figure}[t]
\vspace{-5pt}
    \centering 
     \includegraphics[width=0.24\textwidth]{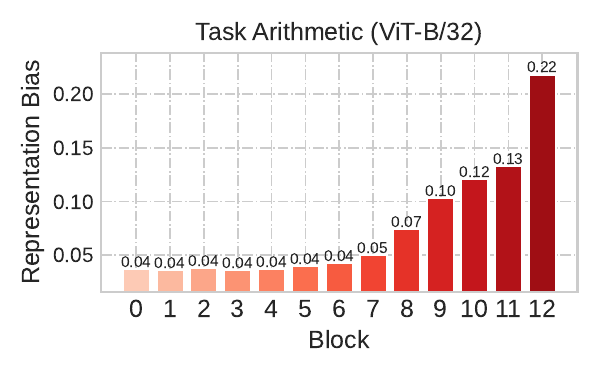}
     \includegraphics[width=0.24\textwidth]{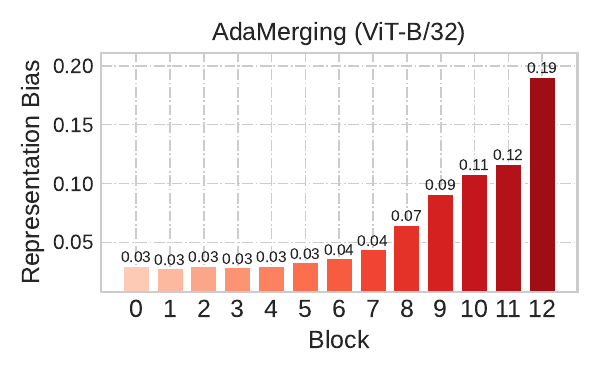}
\vspace{-20pt}
    \caption{An illustration of layer-wise representation bias. The $x$-axis represents the ID (or depth) of the block/layer, while the $y$-axis represents the representation bias of the corresponding layer. The four figures from left to right are combinations of two model merging methods (i.e., Task Arithmetic and AdaMerging) and ViT-B/32 architecture.
    }  
\label{fig:layerwise_bias_tv_vit_b32} 
\vspace{-15pt}
\end{figure}

\textbf{Observation 2}: \textit{The ``representation bias'' problem exists at every layer of the merged model and increases monotonically with the layer's depth.} 
Given that the forward propagation process of a deep neural network model follows a directed acyclic computation graph, the vanilla {\cmethodname} in $\S$\ref{subsec:surgery} measures the representation bias and performs the representation surgery at the \textit{last} layer. However, we argue that representation bias exists at every layer and interacts in complex ways. This leads to systemic bias in the final output that cannot be completely corrected by adjusting only the last layer. 
Specifically, as shown in Fig.~\ref{fig:layerwise_bias_tv_vit_b32}~\footnote{\footnotesize Due to space limitations, Fig.~\ref{fig:layerwise_bias_tv_vit_b32} shows the average layer-wise representation bias across eight tasks. The layer-wise representation bias for each task is depicted in Fig.~\ref{fig:compare_surgery_v1v2_layerwise_bias_tv_vit_b32_appendix} and Fig.~\ref{fig:compare_surgery_v1v2_layerwise_bias_adamerging_vit_b32_appendix} in Appendix~\ref{subsec:analysis_appendix}. Similar trends are observed across all eight tasks.}, 
by extending Definition~\ref{def:representation_bias}, we quantify the representation bias $\small \{b_1^{(t)},\ldots,b_L^{(t)}\}$ of layer $\small \{1,\ldots,L\}$ w.r.t task $t$ following the definition provided in Definition \ref{def:representation_bias}. 
Observations reveal that: 
(i) The representation bias $b_l^{(t)}$ of every layer is non-zero and positive, indicating the presence of representation bias across all layers of the merged model.
(ii) Representation bias monotonically increases with the layer's depth, whereby deeper layers exhibit a larger bias compared to shallower ones. This phenomenon arises from the accumulation of representation bias from shallow to deep layers during forward propagation. These two observations remain consistent across various architectures, model merging methods, and datasets (eight datasets are shown in the Appendix~\ref{subsec:analysis_appendix}), indicating their generality across different settings rather than being specific to particular instances.

\begin{figure*}[t]
    \centering 
    \includegraphics[width=0.16\textwidth]{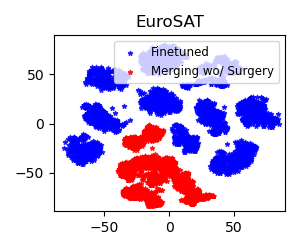}
    \vspace{-2.1pt}
     \includegraphics[width=0.16\textwidth]{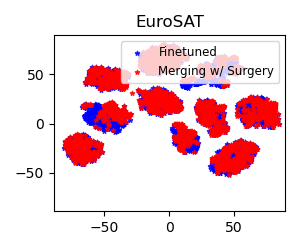}
     \vspace{-2.1pt}
     \includegraphics[width=0.16\textwidth]{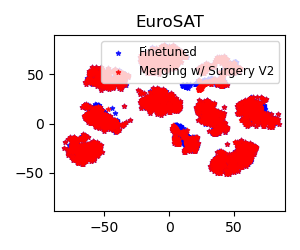}
     \vspace{-2.1pt}
      \includegraphics[width=0.16\textwidth]{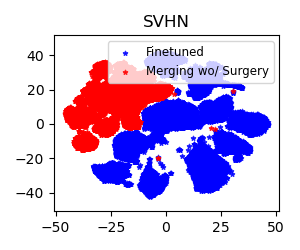}
     \includegraphics[width=0.16\textwidth]{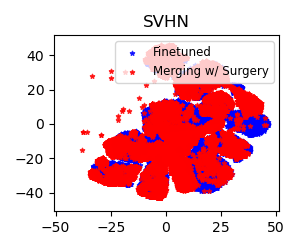}
     \includegraphics[width=0.16\textwidth]{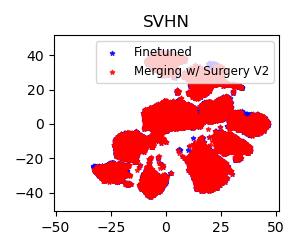}
    \caption{Visualization of the distribution of final-layer representations for individual models (\textcolor{blue}{blue}) and merged models (\textcolor{red}{red}) on EuroSAT and SVHN datasets.
    The \textit{first column} represents the merged model obtained by Task Arithmetic without the representation surgery scheme. The \textit{second column} indicates the use of the {\cmethodname} scheme from $\S$\ref{subsec:surgery}, while the \textit{third column} shows the results using our proposed {\methodname} scheme in $\S$\ref{subsec:ourmethod}. Additional visualizations for more datasets are provided in Fig.~\ref{fig:distribution_tv_vit_b32_part1_appendix}, Fig.~\ref{fig:distribution_tv_vit_b32_part2_appendix}, Fig.~\ref{fig:distribution_adamerging_vit_b32_part1_appendix}, and Fig.~\ref{fig:distribution_adamerging_vit_b32_part2_appendix} of Appendix~\ref{subsec:analysis_appendix}.
    }  
\label{fig:visual_tv_vitb32_resisc45_dtd} 
\vspace{-10pt}
\end{figure*}

Combining the observations mentioned above, we conclude that the \textbf{``representation bias'' problem exists at every layer of the merged model}. Addressing ``representation bias'' only on the last layer as {\cmethodname} ($\S$\ref{subsec:surgery}) is insufficient to fully resolve systemic representation bias, as biases introduced at each layer can accumulate and interact in complex ways.

\subsection{
{\methodname}: Deep Representation Surgery}
\label{subsec:ourmethod}

In this section, to more effectively address the issue of ``representation bias'' in the merged model, we introduce a novel deep representation surgery scheme, referred to as {\methodname} (as shown in Fig.~\ref{fig:method}(d)).

Unlike vanilla {\cmethodname} in $\S$\ref{subsec:surgery}, which accumulates the representation biases from all layers to the \textit{last layer} using a ``{lazy alignment}" strategy, our {\methodname} scheme employs an ``{immediate alignment}" strategy to perform representation surgery at \textit{each layer}. 
Specifically, we first add a lightweight surgery module $\small \Omega^{l}_{\mathbf{W}^{(t)}_l}$ to each layer $l$ w.r.t task $t$ of the merged model $\small \Phi_{\mathbf{\Theta}^{(mtl)}}$. 
Next, we align the representation $\small \mathbf{\hat{Z}}_{l}^{(t), mtl}$ of the merged model $\small \Phi_{\mathbf{\Theta}^{(mtl)}}$ at layer $\small l \in \{1,\ldots,L\}$ with the representation $\small \mathbf{Z}_{l}^{ (t), ind}$ of the individual model $\small \Phi_{\mathbf{\Theta}^{(t)}}$ at the corresponding layer $\small l \in \{1,\ldots,L\}$.
Finally, the overall optimization goal of our {\methodname} method is defined as:
\begin{equation}
\small
\begin{split}
    \min_{\{\mathbf{W}^{(1)}_1,\ldots,\mathbf{W}^{(T)}_L\}} \; \sum_{t=1}^T \sum_{l=1}^L
    \frac{1}{|\mathcal{\hat{D}}|}\; \mathcal{\psi}
    \left(\mathbf{\hat{Z}}_{l}^{(t), mtl}
    , \; \mathbf{Z}_{l}^{ (t), ind}\right),\\
    \text{ where  } \mathbf{\hat{Z}}_{l}^{(t), mtl} = \mathbf{Z}_{l}^{(t), mtl} - \Omega^{l}_{\mathbf{W}^{(t)}_l}(\mathbf{Z}_{l}^{(t), mtl}),
\end{split}
\label{eq:surgery_v2}
\end{equation} 
where $T$ and $L$ denote the number of tasks and layers, respectively. $\mathcal{\psi}(\cdot)$ denotes an arbitrary distance function, which could be $L_1$ loss, mean squared error (MSE), or negative cosine similarity. Detailed comparisons of these different distance functions are provided in Tab.~\ref{tab:lossfun_vitbase32_appendix} of Appendix~\ref{subsec:analysis_appendix}.
In addition, $\small \Omega^{l}_{\mathbf{W}^{(t)}_l}$ is a learnable lightweight module. Without loss of generality, we adopt an adapter-like~\cite{adapter_icml2019} structure as described in $\S$\ref{subsec:surgery}, which is, 
\begin{equation}
\small
\centering
\begin{split}
        \Omega^{l}_{\mathbf{W}^{(t)}_l}(\mathbf{Z}_{l}^{(t), mtl}) = \mathbf{W}^{(t)}_{l,\text{Up}} \texttt{ReLU} (\mathbf{W}^{(t)}_{l,\text{Down}} \cdot \mathbf{Z}_{l}^{(t), mtl}),
\end{split}
\label{eq:surgery_module_v2}
\end{equation}
where $\small \mathbf{W}^{(t)}_l=\{\mathbf{W}^{(t)}_{l,\text{Up}}, \mathbf{W}^{(t)}_{l,\text{Down}}\}$, and $\small \mathbf{W}^{(t)}_{l,\text{Up}} \in \mathbb{R}^{d_l \times r}$ and $\small \mathbf{W}^{(t)}_{l,\text{Down}} \in \mathbb{R}^{r \times d_l}$ are two learnable weights within the {\methodname} module $\small \Omega^{l}_{\mathbf{W}^{(t)}_l}$, and $r$ denotes a hyperparameter, which we set to 16 by default.

However, in the context of model merging, the labeled original training data is unavailable. To optimize $\small \{\mathbf{W}^{(t)}_l\}_{l=1,t=1}^{l=L,t=T}$ in {\methodname}, drawing inspiration from test-time adaptation~\cite{Tent_ICLR2021,TTA_survey2023}, we leverage unlabeled test data. Furthermore, we also verified that the proposed {\methodname} can still work with wild data, although with a slight reduction in performance. Specifically,
(i) In the case of \textit{{\methodname} with Unlabeled Test-time Data}, we set $\small \mathcal{\hat{D}}$ in Eq.~\ref{eq:surgery_v2} to $\small \mathcal{D}_{te}^{(t)}$ as detailed in $\S$\ref{subsec:setting}. 
(ii) In the case of \textit{{\methodname} with the Wild Data}, we utilize some easily accessible public data as $\small \mathcal{\hat{D}}$ in Eq.~\ref{eq:surgery_v2}, e.g., benchmark data like ImageNet~\cite{deng2009imagenet} in the CV domain, and data from {Wikipedia} in the NLP domain. The robust performance under wild data significantly broadens the applicability of our method.

\begin{figure}[t]
\vspace{-5pt}
    \centering 
     \includegraphics[width=0.24\textwidth]{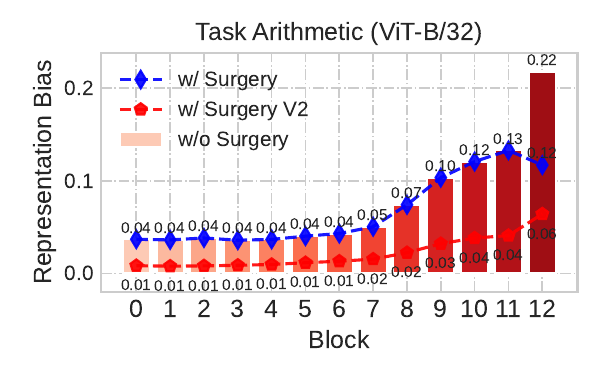}
     \includegraphics[width=0.24\textwidth]{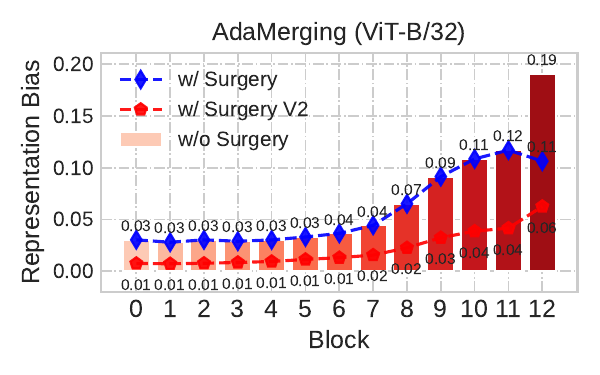}
    \vspace{-15pt}
    \caption{
    Visualization of layer-wise representation bias (lower better). The left and right figures show the two model merging methods of Task Arithmetic and AdaMerging, respectively. In these figures, the red column indicates the representation bias of the merged model without the surgery scheme. The \textit{blue} polyline and \textit{red} polyline represent the representation bias after using the {\cmethodname} scheme from $\S$\ref{subsec:surgery} and the {\methodname} scheme proposed in $\S$\ref{subsec:ourmethod}, respectively. 
    Note that these results are averaged over eight tasks, and the individual layer-wise representation biases for each task shown in Fig.~\ref{fig:compare_surgery_v1v2_layerwise_bias_tv_vit_b32_appendix} and Fig.~\ref{fig:compare_surgery_v1v2_layerwise_bias_adamerging_vit_b32_appendix} in the Appendix~\ref{subsec:analysis_appendix}.
    }  
\label{fig:layerwise_bias_v1v2_vit-b/32} 
\vspace{-15pt}
\end{figure}

\textbf{Discussion}.
When the representation surgery is exclusively applied to the \textit{last layer} as described in Eq.~\ref{eq:surgery_v2}, our {\methodname} reverts to the vanilla Surgery approach in $\S$\ref{subsec:surgery}. Indeed, our {\methodname} demonstrates notably enhanced effectiveness compared to that of {\cmethodname} in $\S$\ref{subsec:surgery}, even when operating at equivalent capacity. 
Subsequently, we provide an analysis from two perspectives as follows:
\begin{itemize}[noitemsep,topsep=0pt,parsep=0pt,partopsep=0pt,leftmargin=*]
    \item \textbf{\textit{Representation Bias}}. 
As depicted in Fig~\ref{fig:layerwise_bias_v1v2_vit-b/32}, we evaluate the block-wise representation bias and observe that the merged model without any surgery scheme has a very large representation bias, while the {\cmethodname} scheme in $\S$\ref{subsec:surgery} corrects the bias only in the last layer. For instance, in the case of the merged model generated through the Task Arithmetic method (left subfigure), the original {\cmethodname} reduces the bias of the last layer from 0.22 to 0.12. However, our {\methodname} not only further reduces the bias of the last layer to 0.06, but also effectively corrects bias from previous layers, for example, the biases of the first 6 layers approximate 0.01. 
    \item \textbf{\textit{Representation Distribution}}. 
We examined the level of overlap between the representation distributions extracted by the merged model and those by the individual model using the t-SNE~\cite{tsne2008} tool. Fig.~\ref{fig:visual_tv_vitb32_resisc45_dtd} illustrates that the overlap between the distributions of {\methodname} (third column) is greater than that of the original {\cmethodname} (second column), whereas the overlap between the distributions without employing any surgery scheme (first column) is the lowest.
\end{itemize}

The aforementioned evidence indicates that {\methodname} holds greater promise in aligning the merged model with the individual expert models, thereby potentially achieving performance levels closer to those of the individual model (or traditional MTL model).

\section{Experiment}
\label{sec:experiment}

\begin{table*}[t]
\centering
\caption{Multi-task performance when merging ViT-B/32 models on eight tasks. $^\ddagger$ indicates that {\cmethodname} uses the same number of parameters as {\methodname}. 
}
\vspace{-5pt}
\label{tab:performance_vitbase32} 
\resizebox{\linewidth}{!}{  
\begin{tabular}{l|cccccccc|cc}
\hline
{Method}  &  {SUN397}  &  {Cars}  &  {RESISC45}  &  {EuroSAT}  &  {SVHN}  &  {GTSRB}  &  {MNIST}  &  {DTD}  & \textbf{Avg.} $\uparrow$  \\
\hline
{Pretrained}  &  {62.3}  &  {59.7}  &  {60.7}  &  {45.5}  &  {31.4}  &  {32.6}  &  {48.5}  &  {43.8} & {48.0}  \\
{Individual}  &  75.3  &  77.7  &  96.1  &  99.7  &  97.5  &  98.7  &  99.7  &  79.4 & 90.5   \\
{Traditional MTL}    &  73.9  &  74.4  &  93.9  & 98.2    &  95.8  &  98.9   &  99.5   & 77.9 & 88.9  \\
\hline
{Weight Averaging} & 65.3  &  63.4  &  71.4  &  71.7  &  64.2  &  52.8  &  87.5  &  50.1  & 65.8 \\
{Fisher Merging}~\citep{FisherMerging_NeurIPS2022}   &  68.6  &  69.2  &  70.7  &  66.4  &  72.9  &  51.1  &  87.9  &  59.9 & 68.3  \\
{RegMean}~\citep{RegMean_ICLR2023}   &  65.3  &  63.5  &  75.6  &  78.6  &  78.1  &  67.4  &  93.7  &  52.0 & 71.8 \\
Task Arithmetic~\citep{TaskArithmetic_ICLR2023}   &55.2 &54.9 &66.7 &78.9 &80.2 & 69.7 &97.3 &50.4 & 69.1 \\
{Ties-Merging}~\citep{TiesMerging_NeurIPS2023}   &  65.0  & 64.4  & 74.8  &  77.4  &  81.2  & 69.3 &  96.5  &  54.5 & 72.9  \\
{AdaMerging}~\cite{AdaMerging_ICLR2024}   &64.5 &68.1 &79.2 &93.8 &87.0 &91.9 &97.5  &59.1 &80.1 \\
Concrete AdaMerging~\cite{tang2023concrete} &  67.8 &70.0 & 87.5 &  96.0 & 91.6 & 96.7 & 98.7 & 63.8 &84.0\\
\hline
\rowcolor{mygray}
{Weight Averaging w/ {\cmethodname}} (Ours) & 67.6 &64.6 &85.8 &96.8 &76.9 &82.9 &97.8 &67.3 &80.0\\
\rowcolor{mygray}
Task Arithmetic w/ {\cmethodname} (Ours)  &63.8 &59.9 &83.3 &97.9 &87.0 &87.0 &98.6 &69.4 &80.9\\
\rowcolor{mygray}
Ties-Merging w/ {\cmethodname} (Ours)  &69.8 &66.1 &87.3 &97.5 &86.7 &87.6 &98.5 &71.6 &83.1\\
\rowcolor{mygray}
{AdaMerging w/ {\cmethodname}} (Ours)  &69.8&71.0 &88.9 &98.1 &91.7 &96.5 &98.8 &73.6 &86.1 \\
\hline
\rowcolor{mygray}
{Weight Averaging w/ {\cmethodname}}$^\ddagger$ (Ours)  &   70.5 &68.2  & 91.9 & 97.8 & 78.8 &97.0  & 98.5 & 77.6 &85.0 \\
\rowcolor{mygray}
Task Arithmetic w/ {\cmethodname}$^\ddagger$ (Ours)  & 66.2  & 64.8 & 91.9 & 98.6 & 89.0 & 97.4 & 98.9 & 78.0 & 85.6 \\
\rowcolor{mygray}
Ties-Merging w/ {\cmethodname}$^\ddagger$ (Ours)   &  71.5 & 69.7 &  92.6 & 98.5 & 88.6 &97.3  &99.0 & 78.7 &87.0 \\
\rowcolor{mygray}
{AdaMerging w/ {\cmethodname}}$^\ddagger$ (Ours)  & 71.6  & 72.3 & 93.6 & 99.0 &92.7  &98.4  & 99.1 & 78.7 & 88.2 \\
\hline
\rowcolor{mygray}
\textbf{Weight Averaging w/ \methodname} (Ours) & 72.1  &67.6  & 94.1 & 99.4 &96.0 & 98.2 & 99.4& 77.0 &88.0 \\
\rowcolor{mygray}
\textbf{Task Arithmetic w/ \methodname} (Ours)  & 73.8   & 67.9 & 94.5 & 99.6  & 96.8 & 98.8 & 99.5 & 78.0 & 88.6 \\
\rowcolor{mygray}
\textbf{Ties-Merging w/ \methodname} (Ours)  & 74.1  & 67.2 & 93.7 & 99.6 & 96.1 & 98.5 & 99.4 & 78.6 & 88.4\\
\rowcolor{mygray}
\textbf{AdaMerging w/ \methodname} (Ours)  & 74.7  & 71.4 & 95.1 & 99.6 & 96.8 & 98.9 & 99.6 & 78.3 & 89.3 \\ 
\hline
\end{tabular}
}
\vspace{-10pt}
\end{table*}

\begin{table*}[t]
\centering
\caption{Multi-task performance when merging ViT-L/14 models on eight tasks. $^\ddagger$ indicates that {\cmethodname} uses the same number of parameters as {\methodname}. 
}
\vspace{-5pt}
\label{tab:performance_vitlarge14} 
\resizebox{\linewidth}{!}{  
\begin{tabular}{l|cccccccc|cc}
\hline
{Method}    &  {SUN397}  &  {Cars}  &  {RESISC45}  &  {EuroSAT}  &  {SVHN}  &  {GTSRB}  &  {MNIST}  &  {DTD}  & \textbf{Avg.} $\uparrow$ \\
\hline
{Pretrained}  &  {66.8}  &  {77.7}  &  {71.0}  &  {59.9}  &  {58.4}  &  {50.5}  &  {76.3}  &  {55.3} & {64.5}   \\
{Individual}   &  82.3  &  92.4  &  97.4  &  100  &  98.1  &  99.2  &  99.7  &  84.1  & 94.2   \\
{Traditional MTL} &  80.8   &  90.6   &   96.3  & 96.3   & 97.6   & 99.1   &  99.6  &  84.4   & 93.5    \\
\hline
{Weight Averaging}    &  72.1  &  81.6  &  82.6  &  91.9  &  78.2  &  70.7  &  97.1  &  62.8 & 79.6 \\
{Fisher Merging}~\citep{FisherMerging_NeurIPS2022}     &  69.2  &  88.6  &  87.5  &  93.5  &  80.6  &  74.8  &  93.3  &  70.0  & 82.2 \\
{RegMean}~\citep{RegMean_ICLR2023}    &  73.3  &  81.8  &  86.1  &  97.0  &  88.0  &  84.2  &  98.5  &  60.8  & 83.7 \\
{Task Arithmetic}~\citep{TaskArithmetic_ICLR2023}  &73.9  &82.1 &86.6 &94.1  &87.9  &86.7  &98.9  &65.6   &84.5 \\
{Ties-Merging}~\citep{TiesMerging_NeurIPS2023}   &  76.5  &  85.0  &  89.3  &  95.7  &  90.3  &  83.3  &  99.0  &  68.8  & 86.0   \\
{AdaMerging}~\cite{AdaMerging_ICLR2024}   &79.0 &90.3 &90.8 & 96.2 &93.4 &98.0  &99.0  &79.9  &90.8 \\
Concrete AdaMerging~\cite{tang2023concrete} &77.8 &91.2 &92.1 &97.0 &94.4 &97.9 &99.0 &79.5 &91.1\\
\hline
\rowcolor{mygray}
{Weight Averaging w/ {\cmethodname}} (Ours) &73.7 &83.9 &92.0 &98.4 &82.4 &86.3 &98.7 &71.9 &85.9 \\
\rowcolor{mygray}
{Task Arithmetic w/ {\cmethodname}} (Ours)  &75.7 &84.4 &93.1 &98.8 &91.3 &93.4 &99.1 &76.1 &89.0\\
\rowcolor{mygray}
{Ties-Merging w/ {\cmethodname}} (Ours)  &76.5 &85.9 &93.7 &99.2 &89.7 &92.0 &99.1 &78.1 &89.3\\
\rowcolor{mygray}
{AdaMerging w/ {\cmethodname}} (Ours)  &80.3 &90.8 &94.3 &98.2 &94.1 & 98.7 &99.2 &82.5 &92.3 \\
\hline
\rowcolor{mygray}
{Weight Averaging w/ {\cmethodname}}$^\ddagger$ (Ours) & 75.6 &  84.7 & 95.6 & 99.1 & 86.3  & 97.3 & 99.2  & 81.4 & 89.9 \\
\rowcolor{mygray}
Task Arithmetic w/ {\cmethodname}$^\ddagger$ (Ours)  &76.7   &85.4  &95.7  & 99.4 & 92.7 & 98.5 & 99.2 &81.0  & 91.1\\
\rowcolor{mygray}
Ties-Merging w/ {\cmethodname}$^\ddagger$ (Ours)   & 77.4  & 87.2 & 95.7 & 99.5 & 91.3 & 98.2 & 99.3 & 81.9 & 91.3 \\
\rowcolor{mygray}
{AdaMerging w/ {\cmethodname}}$^\ddagger$ (Ours)  & 80.4  & 91.1 &  95.5 &  99.2 & 94.9 & 99.0 & 99.3 & 83.2 & 92.8 \\
\hline
\rowcolor{mygray}
\textbf{Weight Averaging w/ \methodname} (Ours) &77.5   & 86.2 & 96.0 & 99.5 &97.8 & 99.1 & 99.6 & 81.3  &92.1 \\
\rowcolor{mygray}
\textbf{Task Arithmetic w/ \methodname} (Ours)  & 80.4  & 88.0 & 96.6 & 99.6 & 97.8 & 99.1 &99.6 & 81.7 & 92.8 \\
\rowcolor{mygray}
\textbf{Ties-Merging w/ \methodname} (Ours)  & 80.0  & 88.7 & 96.3 & 99.7 & 97.9 & 99.0 & 99.6 & 81.5 & 92.8 \\
\rowcolor{mygray}
\rowcolor{mygray}
\textbf{AdaMerging w/ \methodname} (Ours)  &83.8   &91.5  & 96.7 & 99.6 & 97.9 & 99.1 & 99.6 & 84.0 & 94.0 \\
\hline
\end{tabular}
}
\vspace{-5pt}
\end{table*}

This section presents the main results. Due to page limitations, additional experimental results and analysis are placed in the {Appendix}~\ref{subsec:performance_appendix} and {Appendix}~\ref{subsec:analysis_appendix},  respectively.

\subsection{Experimental Setup}
\label{subsec:experi_setup}

The datasets, architectures, and baselines used in this paper are listed as follows. Further details can be found in Appendix~\ref{subsec:baseline_appendix}.

\textbf{Datasets}: 
Our experiments cover eight vision datasets and five text datasets.
Following the setup of Task Arithmetic~\cite{TaskArithmetic_ICLR2023} and AdaMerging~\citep{AdaMerging_ICLR2024}, we treat the following eight datasets as eight tasks to perform model merging in the \textit{CV domain}: SUN397~\citep{xiao2016sun}, Cars~\citep{krause20133d}, RESISC45~\citep{cheng2017remote}, EuroSAT~\citep{helber2019eurosat}, SVHN~\citep{yuval2011reading}, GTSRB~\citep{stallkamp2011german}, MNIST~\citep{lecun1998mnist}, DTD~\citep{cimpoi2014describing}.
In addition, we treat the following five datasets from \cite{zhang2015character,huang2021continual} as five tasks to perform model merging in the \textit{NLP domain}: AGNews (news classification), Yelp (sentiment analysis), Amazon (sentiment analysis), DBPedia (Wikipedia article classification) and Yahoo (questions and answers categorization). 

\textbf{Architectures}:
For vision data, we use three architectures: ViT-B/32, ViT-B/16, and ViT-L/14 from CLIP~\citep{CLIP_ICML2021}. For text data, we use the commonly used BERT~\citep{bert} model.

\textbf{Baselines}: We compared three \textit{non-model merging} methods: Pretrained, Individual, and Traditional MTL models; seven popular \textit{model merging} methods: Weight Averaging, Fisher Merging~\citep{FisherMerging_NeurIPS2022}, RegMean~\citep{RegMean_ICLR2023}, Task Arithmetic~\citep{TaskArithmetic_ICLR2023}, Ties-Merging~\citep{TiesMerging_NeurIPS2023}, AdaMerging~\citep{AdaMerging_ICLR2024}, Concrete AdaMerging~\citep{tang2023concrete}.

\subsection{Performance Comparison}
\label{subsec:performance}

\textbf{Computer Vision Domain}.
As shown in Tabs.~\ref{tab:performance_vitbase32} and ~\ref{tab:performance_vitlarge14}, we merged the models of ViT-B/32 and ViT-L/14 architectures well-trained on eight tasks, respectively. 
The observations are as follows:
(i) The Individual model performs the best but requires separate model parameters for each task. Traditional MTL collaboratively trains a model on the original training data, and the performance is second-best. Pretrained model performs worst due to the lack of task-specific knowledge.
(ii) Advanced model merging methods, such as Fisher Merging, RegMean, Task Arithmetic, Ties-Merging, AdaMerging and Concrete AdaMerging, outperform simple weight averaging. However, due to representation bias, a significant performance gap remains between these methods and the Traditional MTL paradigm.
(iii) When our {\cmethodname} scheme is used in the existing model merging scheme, the performance of the merged model is significantly improved. However, since the representations are aligned only in the last layer, this scheme still cannot reach the level of Traditional MTL most of the time.
(iv) Our {\methodname} scheme enables the merged model's performance to reach the Traditional MTL model's level. For example, when AdaMerging is combined with {\methodname}, the merged model (89.3 in ViT-B/32) even slightly exceeds the traditional MTL (88.9 in ViT-B/32). 

\textbf{Natural Language Processing Domain}.
We merged BERT models trained separately on five distinct text datasets. As shown in Tab.~\ref{tab:bert_performance}, there is a clear performance gap between model merging based MTL (Weight Averaging and Task Arithmetic) and Traditional MTL. After performing our vanilla {\cmethodname}, the performance of Weight Averaging and Task Arithmetic improved from 55.8 and 64.3 to 74.8 and 74.6, respectively, yet it remains below that of Traditional MTL at 75.0. However, after performing our {\methodname}, the performance of the merged model can be further improved to 75.8 and 76.2, which is very close to the performance of the individual model (i.e., 76.2). This demonstrates the effectiveness of our approach in the NLP domain.

\begin{table*}[t]
\small
\centering
\caption{Multi-task performance when merging BERT models on five tasks.}
\vspace{-5pt}
\label{tab:bert_performance} 
\resizebox{\linewidth}{!}{  
\begin{tabular}{l|ccccc|ccccc}
\hline
{Method}  &  {AG News}  &  {Yelp Sentiment}  &  {Amazon Sentiment}  &  {Yahoo Q\&A}   &  {DBPedia Wikipedia}  & \textbf{Avg.} $\uparrow$  \\
\hline
Individual &91.6 &60.4 &57.8 &72.6 &98.8 & 76.2 \\
Traditional MTL  & 90.6 & 59.1 & 55.6 & 71.3 & 98.5 & 75.0  \\
\hline
Weight Averaging  & 79.2 & 49.8 & 45.0 & 50.3 & 55.1 & 55.8 \\
Task Arithmetic~\cite{TaskArithmetic_ICLR2023}  & 82.9 & 55.8 & 48.4 & 53.1 & 81.5 &  64.3  \\
\hline
\rowcolor{mygray}
{Weight Averaging w/ {\cmethodname}} (Ours) &90.5 &58.5 &55.0 &71.3 &98.8 &74.8 \\
\rowcolor{mygray}
{Task Arithmetic w/ {\cmethodname}} (Ours)  &89.9 &58.0 & 55.8 &71.0 & 98.4 &74.6 \\
\hline
\rowcolor{mygray}
\textbf{Weight Averaging w/ \methodname} (Ours) &90.5 & 60.3& 57.4& 72.0& 98.8&75.8 \\
\rowcolor{mygray}
\textbf{Task Arithmetic w/ \methodname} (Ours) &91.2 & 60.7 &57.5 &72.8 &98.7 & 76.2\\
\hline
\end{tabular}
}
\vspace{-5pt}
\end{table*}

\subsection{Experimental Analysis}
\label{subsec:analysis}

\textbf{{\cmethodname} and {\methodname} with the Wild Data}. 
In cases where unlabeled test data is insufficient for optimizing surgery modules (Eq.~\ref{eq:surgery_v1} in $\S$\ref{subsec:surgery} and Eq.~\ref{eq:surgery_v2} in $\S$\ref{subsec:ourmethod}), we investigated the possibility of utilizing public/auxiliary data as an alternative.
For example, in the upper section of Tab.~\ref{tab:bert_performance_wild}, we merge the models trained on the four datasets of AG, Yelp, Amazon, and Yahoo, and use Wikipedia data for surgery. {\methodname} still achieves the same result as the Traditional MTL, which is 69.3, while the original {\cmethodname} only has 66.9. 
As shown in the lower section of Tab.~\ref{tab:bert_performance_wild}, it is similar to using AG News as wild data to merge the corresponding models of Yelp, Amazon, Yahoo, and DBPedia. 
{These results indicate that even with wild data, {\methodname} also demonstrates the wide applicability and potential to calibrate the merged model without requiring unlabeled test data.}

\begin{table*}[t]
\small
\centering
\caption{Multi-task performance when merging BERT models on four tasks in the wild data.}
\vspace{-5pt}
\label{tab:bert_performance_wild} 
\resizebox{\linewidth}{!}{  
\begin{tabular}{l|c|cccc|cccccc}
\hline
{Method}  & {Wild Data}   &  {AG News}  &  {Yelp Sentiment}  &  {Amazon Sentiment}  &  {Yahoo Q\&A}  & \textbf{Avg.} $\uparrow$  \\
\hline
Traditional MTL  &  -   & 90.0    & 59.5    &  56.9 & 70.8  & 69.3  \\
{Task Arithmetic}~\citep{TaskArithmetic_ICLR2023} &  -   &  81.4     & 57.3  &  51.4 &  45.8 & 59.0  \\
\rowcolor{mygray}
Task Arithmetic w/ {\cmethodname} (Ours)  & DBPedia Wikipedia     &  88.8     &  57.9 & 53.9  & 67.2 &  66.9  \\
\rowcolor{mygray}
\textbf{Task Arithmetic w/ \methodname} (Ours)  & DBPedia Wikipedia   & 90.6     & 60.1  &  55.5 &  71.1 & 69.3  \\
\hline
\hline
{Method}  & {Wild Data}   &  {Yelp Sentiment}  &  {Amazon Sentiment}  &  {Yahoo Q\&A}  & {DBPedia Wikipedia} & \textbf{Avg.} $\uparrow$  \\
\hline
Traditional MTL  &  -   &  59.7   &  55.6   & 71.7  & 98.6  & 71.4  \\
{Task Arithmetic}~\citep{TaskArithmetic_ICLR2023} &  -   &  57.5     & 54.3  & 54.0  & 94.3  & 65.0  \\
\rowcolor{mygray}
Task Arithmetic w/ {\cmethodname} (Ours)  & AG News     & 55.0      &  55.5 & 66.4  & 97.6 &  68.6  \\
\rowcolor{mygray}
\textbf{Task Arithmetic w/ \methodname} (Ours)  &AG News   &  59.8     &   57.4 & 71.1  &  98.8 &  71.8 \\
\hline
\end{tabular}
}
\vspace{-5pt}
\end{table*}

\begin{figure*}[t]
\vspace{-5pt}
    \centering 
     \includegraphics[width=0.245\textwidth]{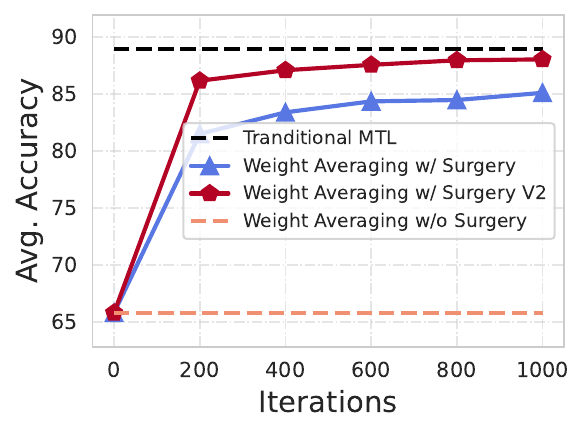}
     \includegraphics[width=0.245\textwidth]{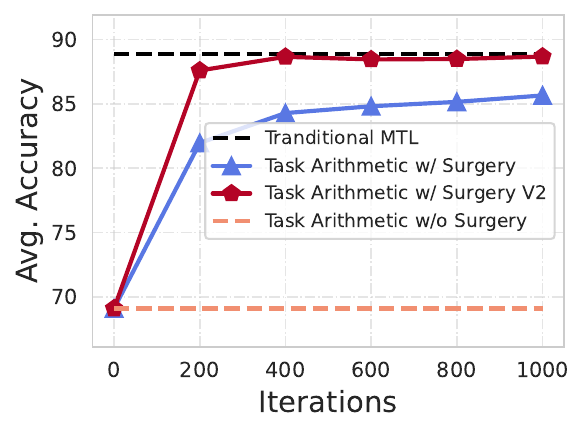}
      \includegraphics[width=0.245\textwidth]{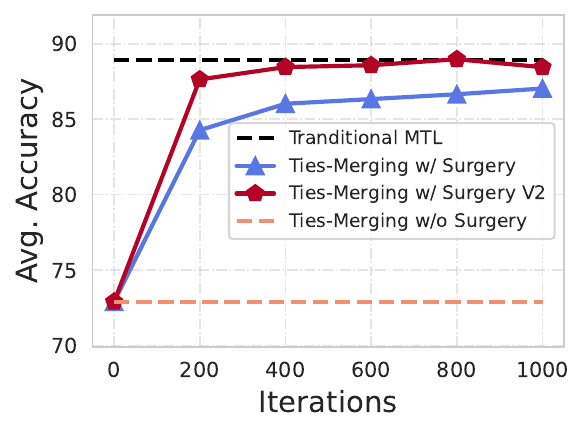}
       \includegraphics[width=0.245\textwidth]{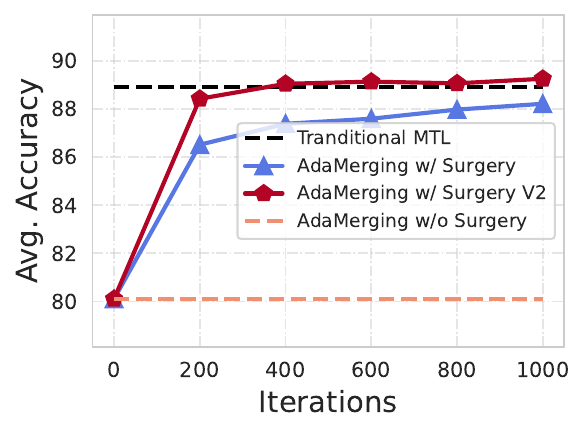}
    \vspace{-15pt}
    \caption{
    An illustration of how the merged model's performance varies with the number of iterations when merging eight ViT-B/32 models. From left to right, the plots represent model merging solutions based on Weight Averaging, Task Arithmetic, Ties-Merging, and AdaMerging.
    }  
\label{fig:iter_acc_vit-b32} 
\vspace{-15pt}
\end{figure*}

\textbf{Performance w.r.t. Iteration}. As shown in Fig.~\ref{fig:iter_acc_vit-b32}, we show the performance changes during iterations of our {\cmethodname} and our {\methodname}. For a fair comparison, the two maintain consistent representation capacity. We can observe that under the four model merging methods (Weight Averaging, Task Arithmetic, Ties-Merging, and AdaMerging), using our {\methodname} achieves higher final performance than the vanilla {\cmethodname}. Even the performance of {\methodname} in the 200-th iteration exceeds the performance of {\cmethodname} in 1000 iterations. This means our {\methodname} is more efficient. 

\section{Conclusion and Future Works}
\label{sec:conclusion}

In this paper, we first demonstrate that state-of-the-art model merging methods are affected by representation bias. To address this issue, we propose a representation surgery approach (i.e., {\cmethodname}) that aligns the merged model's representations with those of the independent expert model at the final layer, thereby mitigating the bias. However, the initial {\cmethodname} scheme still leaves a performance gap. Further analysis reveals that representation bias occurs across all layers of the merged model. To effectively alleviate this, we introduce a deep representation surgery scheme, termed {\methodname}, which significantly reduces the representation bias.
Experiments across CV and NLP tasks, merging methods, and architectures show that our solution matches traditional MTL performance.
There are several directions for future research in our work: (i) developing a representation surgery strategy that eliminates the need for training, (ii) investigating model merging with different initializations or architectures; and (iii) extending the proposed representation surgery to a broader range of model merging scenarios.

{
\small
\bibliographystyle{IEEEtran}
\bibliography{main}
}

\clearpage

\noindent
\textbf{Appendix Overview}. 
The content of this appendix is organized as follows:

{
\hypersetup{linkcolor=black}
\startcontents[sections] 
\printcontents[sections]{}{}{\setcounter{tocdepth}{2}} 
\vskip 0.2in
\hrule
}

\section{Related Works}
\label{sec:relatedworks_appendix}

In this section, we discuss related work from two directions: model merging for multi-task learning in $\S$\ref{subsec:relatedwork_modelmerging_appendix} and traditional multi-task learning in $\S$\ref{subsec:relatedwork_mtl_appendix}.

\subsection{Model Merging for Multi-Task Learning}
\label{subsec:relatedwork_modelmerging_appendix}

Recent studies in the machine learning community have attempted to achieve MTL using model merging techniques~\cite{modelfusion_survey2023,ZipIt_ICLR2023,akiba2024evolutionary,merging_gradientmarching_iclr2024,Survery_ModelMerging_2024}. However, the performance of simple parameter averaging~\cite{izmailov2018averaging,mcmahan2017communication} degrades significantly as the number of tasks increases, resulting in a substantial gap between its performance and that of traditional MTL model.
Many recent works have made various attempts to fill this gap. We divide model merging into two stages: before and during merging. 

(i) The main concern \textbf{before merging} is how to provide more favorable preconditions for model merging, such as linearization or orthogonalization. Specifically, some works~\cite{TangentSpace_NeurIPS2023,liu2024tangent} independently fine-tune each task in the Tangent space~\cite{NTK_NeurIPS2018} of the pre-trained model and demonstrate that this helps disentangle the weight space from the input space, leading to better model merging. Similarly, Linearization-LoRA~\cite{LinearizationLoRA_ICLR2024} linearly fine-tunes some LoRA modules~\cite{lora_iclr2022} in Tangent space.
In addition, Task Arithmetic~\cite{TaskArithmetic_ICLR2023} pointed out that the orthogonality between task vectors is one of the conditions for successful model merging. From the perspective of loss landscape, OTFusion~\cite{fusionvia_optimaltransport_neurips2020}, GitReBasin~\cite{GitReBasin_ICLR2023} and REPAIR~\cite{REPAIR_ICLR2023} rearrange the neurons in each layer to facilitate weight interpolation. 

(ii) The main focus \textbf{during merging} is how to mitigate interference and conflicts between models~\cite{TaskArithmetic_ICLR2023,wu2023pi,TiesMerging_NeurIPS2023,DARE_Arxiv2023,RegMean_ICLR2023,rewardedsoups_neurips2023,pem_neurIPS2023,daheim2024model,AdaMerging_ICLR2024,Survery_ModelMerging_2024}. 
The \textit{first type} of method explores how to better weigh and combine multiple models~\cite{FisherMerging_NeurIPS2022,zhou2024metagpt,akiba2024evolutionary}. For example, Fisher-Merging~\cite{FisherMerging_NeurIPS2022} performs weighted merging utilizing the importance of each parameter through the Fisher information matrix~\cite{fisher1922mathematical}. RegMean~\cite{RegMean_ICLR2023} reweights and linearly combines rows in weight matrices based on statistics from training data.
AdaMerging~\cite{AdaMerging_ICLR2024} leverages unlabeled test data to automatically learn a set of task-level or layer-level model merging coefficients.
The \textit{second type} of method explores how to merge models in sparse subspaces to reduce interference~\cite{panigrahi2023task,du2024NeurIPSparameter,localizing_ICML2024,zimmer2024ICLRsparse,deep2024della,huang2024emr,davari2023model}. For example, Ties-Merging~\cite{TiesMerging_NeurIPS2023} removes the smaller magnitude parameters and eliminates the issue of parameter sign conflicts during model merging. DARE~\cite{DARE_Arxiv2023} removes a large number of useless neuron updates and then scales the neurons for merging. 
Concrete~\cite{tang2023concrete} finds a shared subspace between multiple tasks for model merging. PCB-Merging~\cite{du2024NeurIPSparameter} utilizes intra-balancing to measure parameter importance in individual tasks and inter-balancing to assess parameter similarity between different tasks.
The \textit{third type} of method dynamically merges multiple expert modules during inference~\cite{tang2024merging,muqeeth2024learning,li2024ICLRmerge,lu2024twin,yadav2024survey}. For example, WEMoE~\cite{tang2024merging} dynamically merges linear layers by routing, and static merges nonlinear layers. It should be noted that the dynamic merging method requires additional maintenance of more model parameters than the first two categories of methods, and it also reduces the inference efficiency.

While existing methods predominantly concentrate on merging in weight space, they often neglect a crucial concern stemming from weight merging--the \textit{representation bias}. A substantial disparity emerges in the \textit{representation space} between the merged model and individually trained expert models. In contrast, our surgery method addresses this gap, aiming to minimize the representation discrepancy. Moreover, our approach operates in the \textit{representation space}, offering a complementary and orthogonal perspective to traditional weight-space merging methods. Consequently, our method can be seamlessly integrated with them.

\subsection{Traditional Multi-Task Learning}
\label{subsec:relatedwork_mtl_appendix}

MTL uses one model to accommodate information from multiple tasks, which is very efficient in terms of computational efficiency~\cite{SharedBottom_1997,mtlsurvey_tpami2021}. However, 
MTL usually faces negative transfer~\cite{SharedBottom_1997,mtlsurvey_tpami2021,negative_transfer_survey_2022}.
Existing work mainly solves the negative transfer problem from two directions: architecture and optimization. Specifically,
(i) The classic SharedBottom~\cite{SharedBottom_1997} backbone performs poorly when tasks are not highly correlated. Advanced \textit{{architectures}} primarily alleviate the phenomenon of negative transfer through modularization~\cite{mmoe_kdd2018,ple_recsys2020}, sparsification~\cite{Adashare_NeurIPS2020,dwa_cvpr2019}, and soft sharing~\cite{CrossStitch_CVPR2016,mtlnas_CVPR2020} of the backbone. 
(ii) Other work alleviates task interference from an \textit{{optimization}} perspective. For example, some works~\cite{uncertaintyweighting_cvpr2018,mtlasmooSenerK18_neurips2018,dwa_cvpr2019,revisitingScalarization_NeurIPS2023} try to optimize the weight for each task loss.  Some works~\cite{graddrop_neurips2020,PCGrad_NeurIPS2020,CAGrad_NeurIPS2021,wang2021gradientVaccine,nashmtl_ICML2022} resolve multi-task gradient direction or sign conflicts. Other works~\cite{gradnorm_icml2018,imtl_iclr2021,MetaBalanceWWW2022,adatask_aaai2023} try to eliminate the dominance of the learning rate or gradient. In particular, KD4MTL~\cite{kd4mtl2020} uses the model trained on a single task as an additional supervised loss to guide the re-training of the MTL model, thereby alleviating the difficulty and magnitude imbalance of the loss in the MTL joint training.

Distinct from these existing traditional MTL approaches that concentrate on loss weight or gradient space to tackle the negative transfer issue, our method improves MTL performance by addressing the representation bias problem within the representation space in model merging based MTL, providing a novel and orthogonal perspective.

\section{Experimental Details}
\label{sec:experimental_details_appendix}
\subsection{Datasets}
Following Task Arithmetic~\cite{TaskArithmetic_ICLR2023}, Ties-Merging~\cite{TiesMerging_NeurIPS2023} and AdaMerging~\cite{AdaMerging_ICLR2024}, we merge the models trained on the following eight datasets.
\begin{itemize}[noitemsep,topsep=0pt,parsep=0pt,partopsep=0pt]
    \item \textbf{SUN397}~\citep{xiao2016sun}: The database is a benchmark dataset for Scene Understanding (SUN) and contains a total of 108,753 images from 397 classes, where each class contains a different number of images, but each class has at least 100 images.
    \item \textbf{Cars}~\citep{krause20133d}: Stanford Cars is a dataset used for fine-grained recognition in the field of computer vision. It contains images of 196 car classes with a total of 16,185 images. The images of each class are divided roughly 1:1 into training set and test set. 
    \item \textbf{RESISC45}~\citep{cheng2017remote}: The RESISC45 dataset is a publicly available benchmark for scene classification in remote sensing images. It contains 45 scene classes, and each class contains 700 images (each image resolution is 256$\times$256), for a total of about 31,500 images.
    \item \textbf{EuroSAT}~\citep{helber2019eurosat}: EuroSAT is a Sentinel-2-based satellite image dataset primarily used to classify land use in geospatial imagery and contains 27,000 labeled and geo-referenced images in 10 classes.
    \item \textbf{SVHN}~\citep{yuval2011reading}: SVHN is a real image dataset containing 10 classes of color house number images, SVHN is very similar in style to MNIST~\citep{lecun1998mnist} (handwritten digits in grayscale images), but it contains a larger number of images (more than 600,000 digital images).
    \item \textbf{GTSRB}~\citep{stallkamp2011german}: The German Traffic Sign Recognition Benchmark (GTSRB) contains images with different lighting conditions and rich backgrounds. These images are classified into 43 classes of traffic signs, totaling more than 50,000 images.
    \item \textbf{MNIST}~\citep{lecun1998mnist}: MNIST is a large database of handwritten digits in 10 classes that is one of the most famous datasets in machine learning. It contains 60,000 training images and 10,000 test images, each of which is 28x28 pixels.
    \item \textbf{DTD}~\citep{cimpoi2014describing}: The Describable Textures Dataset (DTD) contains 5,640 real labeled texture images, divided into 47 classes, each with about 120 images.
\end{itemize}

\subsection{Baselines} 
\label{subsec:baseline_appendix}

The baselines we compare include three non-model merging methods, and seven representative model merging methods:

\textbf{(i) Non-model Merging Methods}:
\begin{itemize}[noitemsep,topsep=0pt,parsep=0pt,partopsep=0pt,leftmargin=*]
     \item \textbf{\textit{Pretrained}} directly uses the pre-trained model to test multiple downstream tasks. The performance of this method is generally less than acceptable due to the lack of task-related knowledge.
     \item \textbf{\textit{Individual}} involves full fine-tuning of the pretrained model with task-specific training data and often yields superior performance. However, it requires maintaining a copy of model parameters for each task, which is very costly when the number of tasks is large.
     \item \textbf{\textit{Traditional MTL}} collects data from multiple tasks and collaboratively trains a multi-task model. Due to potential task interference, the performance of a multi-task model is usually slightly lower than that of individual models, but the advantage is that it only has a single model.
\end{itemize}

\textbf{(ii) Model Merging Methods}:
\begin{itemize}[noitemsep,topsep=0pt,parsep=0pt,partopsep=0pt,leftmargin=*]
     \item \textbf{\textit{Weight Averaging}} is the simplest model merging method, which directly adds the parameters of well-trained multiple models according to average values.
     \item \textbf{\textit{Fisher Merging}}~\citep{FisherMerging_NeurIPS2022} measures the importance of each parameter based on the Fisher information matrix~\citep{fisher1922mathematical} and merges models based on that importance.
     \item \textbf{\textit{RegMean}}~\citep{RegMean_ICLR2023} utilizes pre-computed inner product matrices of inputs at each layer from the original training data to conduct a linear combination of the model's parameters.
     \item \textbf{\textit{Task Arithmetic}}~\citep{TaskArithmetic_ICLR2023} defines the parameter difference between the fine-tuned model and the pre-trained model as a `task vector', and merges multiple task vectors into the pre-trained model based on grid search coefficients for model merging.
     \item \textbf{\textit{Ties-Merging}}~\citep{TiesMerging_NeurIPS2023} eliminates sign conflicts between parameters of task vectors based on Task Arithmetic~\citep{TaskArithmetic_ICLR2023}, thereby effectively alleviating interference during model merging.
     \item \textbf{\textit{AdaMerging}}~\citep{AdaMerging_ICLR2024} designed a layer-wise model merging strategy, which assigns a merging coefficient to each layer of the model to be merged and optimizes these coefficients by entropy minimization.
     \item \textbf{\textit{Concrete AdaMerging}}~\citep{tang2023concrete} performs model merging in the shared subspace of multiple models. Similar to AdaMerging~\citep{AdaMerging_ICLR2024}, it employs entropy minimization to identify the shared subspace.
\end{itemize}

\textbf{(iii) Our Representation Surgery Methods}:
Our {\cmethodname} (in $\S$\ref{subsec:surgery}) is orthogonal to the model merging methods in (ii). It points out that the merged model has `representation bias' issue, and proposes to align the representation of the \textit{last layer} of the individual model and the merged model to alleviate the bias. It contains the following four variants:
\begin{itemize}[noitemsep,topsep=0pt,parsep=0pt,partopsep=0pt,leftmargin=*]
     \item \textbf{\textit{Weight Averaging w/ {\cmethodname}}} performs representation surgery on the merged model by the Weight Averaging method.
     \item \textbf{\textit{Task Arithmetic w/ {\cmethodname}}} carries out the representation surgery on the merged model via the Task Arithmetic~\citep{TaskArithmetic_ICLR2023} method.
     \item \textbf{\textit{Ties-Merging w/ {\cmethodname}}} executes the surgery in the combined model through the Ties-Merging~\citep{TiesMerging_NeurIPS2023} method.
     \item \textbf{\textit{AdaMerging w/ {\cmethodname}}} conducts the representation surgery within the model that has been merged using the AdaMerging~\citep{AdaMerging_ICLR2024} method.
\end{itemize}

\textbf{(iv) Our Deep Representation Surgery Methods}:
Our {\methodname} (in $\S$\ref{subsec:ourmethod}), like vanilla {\cmethodname} (in $\S$\ref{subsec:surgery}), is also orthogonal to the model merging methods in (ii). In contrast to vanilla {\cmethodname}, we point out that `representation bias' exists in every layer of the merged model, and therefore propose to perform representation surgery at \textit{each layer} to align the representations of the individual model and the merged model. Furthermore, we incorporated the following four variants for experimental evaluation:
\begin{itemize}[noitemsep,topsep=0pt,parsep=0pt,partopsep=0pt,leftmargin=*]
     \item \textbf{\textit{Weight Averaging w/ SurgeryV2}} performs {\methodname} to correct the representation bias of the Weight Averaging merged model. 
     \item \textbf{\textit{Task Arithmetic w/ SurgeryV2}} conducts a {\methodname} on each layer of the merged model via Task Arithmetic~\citep{TaskArithmetic_ICLR2023} to align it with the individual model.
     \item \textbf{\textit{Ties-Merging w/ SurgeryV2}} first obtains a merged model through Ties-Merging~\citep{TiesMerging_NeurIPS2023}, and then executes the {\methodname} scheme proposed in this paper on it.
     \item \textbf{\textit{AdaMerging w/ SurgeryV2}} performs {\methodname} on merged model by the AdaMerging~\citep{AdaMerging_ICLR2024}.
\end{itemize}

\subsection{Implementation Details}
A critical hyperparameter in our {\cmethodname} or {\methodname} methods is the rank $r$ in Eq.~\ref{eq:surgery} ($\S$\ref{subsec:surgery}) or Eq.~\ref{eq:surgery_module_v2} ($\S$\ref{subsec:ourmethod}), which demonstrates the capacity of the representation surgery module. For the ViT-B/32 and ViT-B/16 architectures, we default it to 16; for the ViT-L/14 and BERT architectures, it defaults to 4. The results in Tab.~\ref{tab:surgeryv2_rank_appendix} show that our method is very robust to the setting of rank (e.g., $r \in \{4,6,16,32,64\}$), and a smaller rank can also achieve satisfactory performance.

To optimize the parameters in the {\cmethodname} or {\methodname} module, we employ public data (e.g., ImageNet~\citep{deng2009imagenet}) or unlabeled test data. We use the Adam optimizer~\citep{adam_2014} with a learning rate of 0.001 and a momentum of (0.9, 0.999) to update parameters. The batch size is set to 16, and the number of iterations is set to 1,000 by default. In addition, the loss function $\mathcal{\psi}(\cdot)$ in Eq.~\ref{eq:surgery_v2} of {\methodname} (or Eq.~\ref{eq:surgery_v1} of {\cmethodname}) is set to the $L_1$ function by default. 

We report the accuracy for each task, as well as the average accuracy (i.e., \textbf{Avg.}) across the eight vision tasks or five text tasks. Higher accuracy means better performance of the merged model.


\begin{table*}[t]
\centering
\caption{Multi-task performance when merging ViT-B/16 models on eight tasks. $^\ddagger$ indicates that {\cmethodname} uses the same number of parameters as {\methodname}. }
\vspace{-5pt}
\label{tab:performance_vitbase16_appendix} 
\resizebox{\linewidth}{!}{  
\begin{tabular}{l|cccccccc|cc}
\hline
{Method}   &  {SUN397}  &  {Cars}  &  {RESISC45}  &  {EuroSAT}  &  {SVHN}  &  {GTSRB}  &  {MNIST}  &  {DTD}  & \textbf{Avg.} $\uparrow$ \\
\hline
Pretrained  &63.8 &64.6 &65.7 &54.5 & 52.0&43.3 & 51.7&45.1 &55.0\\
Individual  &81.8 &86.8 &96.9 &99.7 &97.8 &99.1 &99.7 &82.0 &92.9\\
\hline
Weight Averaging  &67.7 &70.0 &75.3 &79.5 &74.9 &60.1 &94.4 &43.8 &70.7\\
Fisher-Merging~\citep{FisherMerging_NeurIPS2022}  &68.5 &69.9 &75.2 &80.4 &73.2 &61.2 &94.5 &50.7 &71.7\\
RegMean~\citep{RegMean_ICLR2023} &69.1 &71.6 &77.6 &88.8 &83.7 &70.2 &96.9 &54.6 &76.6\\
{Task Arithmetic}~\citep{TaskArithmetic_ICLR2023} &61.1 &65.9 &74.0 &76.2 &88.0 &73.9 &98.4  &53.0 &73.8 \\
{Ties-Merging}~\citep{TiesMerging_NeurIPS2023} &69.1 &72.5 &80.5 &84.0 &85.0 &71.5 &98.1  &54.9 &77.0 \\
{AdaMerging}~\cite{AdaMerging_ICLR2024})  &70.2 &80.7 &81.6 &94.8 &91.6 &95.8 &98.5  & 66.2 &84.9 \\
\hline

\rowcolor{mygray}
{Weight Averaging w/ {\cmethodname}} (Ours)   &70.3 &72.4 &88.8 &97.6 &82.0 & 83.1 &98.1 &68.5 &82.6 \\

\rowcolor{mygray}
{Task Arithmetic w/ {\cmethodname}} (Ours)    &68.3 &72.3 &88.7 & 97.7 &91.0 &89.5 &98.9 &72.9 &84.9\\

\rowcolor{mygray}
{Ties-Merging w/ {\cmethodname}} (Ours)    &73.0 &76.2 &90.7 &98.1 & 89.7 &86.7 &98.7 &75.2 &86.0\\

\rowcolor{mygray}
{AdaMerging w/ {\cmethodname}} (Ours)   &73.6 &81.5 &90.4 &98.5 &93.2 &97.4 & 98.9 &77.0 &88.8\\
\hline

\rowcolor{mygray}
{Weight Averaging w/ {\cmethodname}}$^\ddagger$ (Ours)   &73.6   & 78.3 & 93.9 & 98.6 & 84.5& 96.8 &99.0 & 81.2 & 88.2\\

\rowcolor{mygray}
Task Arithmetic w/ {\cmethodname}$^\ddagger$ (Ours)   & 71.2  &76.8  &  93.6& 99.0 &92.1 & 97.8 & 99.2& 80.4 & 88.8\\

\rowcolor{mygray}
Ties-Merging w/ {\cmethodname}$^\ddagger$ (Ours)   & 74.9  & 80.1 & 94.5 & 98.8  &90.9 &97.8  &99.1 & 81.7 &89.7 \\

\rowcolor{mygray}
{AdaMerging w/ {\cmethodname}}$^\ddagger$ (Ours)   &75.0   & 83.2 &94.2  & 99.2 & 93.9 & 98.3 &99.1 & 81.9 & 90.6 \\
\hline
\rowcolor{mygray}
\textbf{Weight Averaging w/ \methodname} (Ours) & 75.2  & 80.2 & 95.3 & 99.4 & 97.2 & 98.7 &99.6 & 80.5 &90.7 \\
\rowcolor{mygray}
\textbf{Task Arithmetic w/ \methodname} (Ours)  & 77.5  & 80.8 & 95.7 &99.5  & 97.4& 98.8 &99.5 &80.3  & 91.2\\
\rowcolor{mygray}
\textbf{Ties-Merging w/ \methodname} (Ours)  & 77.7  & 81.3 & 96.0 & 99.6 & 97.4 & 99.0 & 99.5 & 80.0 & 91.3 \\
\rowcolor{mygray}
\textbf{AdaMerging w/ \methodname} (Ours)  & 78.6  & 83.9 &95.6  & 99.6 & 97.3 & 99.1 &99.6 & 81.5 &91.9 \\ 
\hline
\end{tabular}
}
\vspace{-5pt}
\end{table*}

\begin{table*}[t]
\centering
\caption{Impact of the amount of {online test data} on model performance in {\methodname} module.}
\vspace{-5pt}
\label{tab:online_testset_ratio_batchsize1_appendix} 
\resizebox{\linewidth}{!}{  
\begin{tabular}{l|c|cccccccc|c}
\hline
{Method}  & Available Test Set &  {SUN397}  &  {Cars}  &  {RESISC45}  &  {EuroSAT}  &  {SVHN}  &  {GTSRB}  &  {MNIST}  &  {DTD}  & \textbf{Avg.} $\uparrow$  \\
\hline
{Traditional MTL}   & -  &  73.9  &  74.4  &  93.9  & 98.2    &  95.8  &  98.9   &  99.5   & 77.9 & 88.9  \\
{Task Arithmetic}~\citep{TaskArithmetic_ICLR2023}  & -  &55.2 &54.9 &66.7 &78.9 &80.2 & 69.7 &97.3 &50.4 & 69.1 \\
\hline
\rowcolor{mygray}
\textbf{Task Arithmetic w/ \methodname} (Ours) & 1\%   & 68.8   &   61.7    &  76.6   &  84.0   &  94.6  & 88.2   &  98.5 & 51.9 & 78.0 \\
\rowcolor{mygray}
\textbf{Task Arithmetic w/ \methodname} (Ours) & 10\%  & 72.5   &   64.2    &  89.2   &  95.8   &  96.8  & 96.9   & 99.3   &59.8  & 84.3 \\
\rowcolor{mygray}
\textbf{Task Arithmetic w/ \methodname} (Ours) & 20\%  & 73.9   &   66.5    &  92.3   &  98.2   & 97.0   &  97.9  & 99.5  &64.4  & 86.2 \\
\rowcolor{mygray}
\textbf{Task Arithmetic w/ \methodname} (Ours) & 30\%  & 74.5   &   68.4    &  92.7   &  98.4   & 97.0   &  98.4  & 99.6  &66.8  & 86.9 \\
\rowcolor{mygray}
\textbf{Task Arithmetic w/ \methodname} (Ours) & 40\%  & 74.8   &   70.5    &  93.6   &  99.2   &  97.1  & 98.5    & 99.6  &68.8  & 87.7 \\
\rowcolor{mygray}
\textbf{Task Arithmetic w/ \methodname} (Ours) & 50\%  & 75.3   &   70.3    &  94.3   &  99.3   &  97.0  & 98.4   & 99.5  & 69.7 & 87.9 \\
\rowcolor{mygray}
\textbf{Task Arithmetic w/ \methodname} (Ours) & 60\%  & 75.3   &   70.0    &  94.1   &  99.3   & 97.1   & 98.6   & 99.6  & 71.8 & 88.2 \\
\rowcolor{mygray}
\textbf{Task Arithmetic w/ \methodname} (Ours) & 70\%  & 75.7   &   71.1    &  94.4   &  99.6   &  97.3  & 98.5   & 99.6  &72.4  & 88.5 \\
\rowcolor{mygray}
\textbf{Task Arithmetic w/ \methodname} (Ours) & 80\%  & 75.7   &   70.7    &  94.5   &  99.3  &  97.0  &  98.6  & 99.6  & 74.5 & 88.7 \\
\rowcolor{mygray}
\textbf{Task Arithmetic w/ \methodname} (Ours) & 90\%  & 75.7   &   71.8    &  94.6   &  99.5   & 97.2   & 98.7   & 99.6  & 74.3 & 88.9 \\
\rowcolor{mygray}
\textbf{Task Arithmetic w/ \methodname} (Ours) & 100\% & 76.0   &   70.9    &  94.5   &  99.6   & 97.2   & 98.9   & 99.5  & 75.1 & 88.9 \\
\hline
\end{tabular}
}
\end{table*}

\section{Experimental Results and Analysis}

In this section, we provide additional results and analyses that are omitted from the main text due to space constraints.

\subsection{Performance Comparison}
\label{subsec:performance_appendix}

In Tab.~\ref{tab:performance_vitbase32} and Tab.~\ref{tab:performance_vitlarge14} of the main text ($\S$\ref{subsec:performance}), we show the performance of merging the ViT-B/32 and ViT-L/14 architectures. In Tab.~\ref{tab:performance_vitbase16_appendix} below, we further illustrate the results of merging ViT-B/16 models trained on eight vision tasks. The following significant and consistent phenomena can be observed:
(i) Advanced model merging methods (e.g., Ties-Merging, AdaMerging) all achieve better performance than simple Weight Averaging.
(ii) After using our {\cmethodname} to correct the representation bias problem of the merged model, the performance of the merged model has been significantly improved, especially under the optimal surgery capacity (results with ${\ddagger}$). 
(iii) Our {\methodname} achieves the highest performance by resolving the layer-wise representation bias.

\subsection{SurgeryV2 in Online Setting}
Under default settings, we rely on unlabeled test data to optimize parameters in {\methodname}. In real-world scenarios, unlabeled test data may arrive in a streaming manner. Therefore, we evaluated the performance changes of {\methodname} in an online setting. Specifically, as shown in Tab.~\ref{tab:online_testset_ratio_batchsize1_appendix}, we tested the performance of {\methodname} when $\small \{1\%, 10\%, 20\%,\ldots,100\%\}$ of test data is visible, with each sample being visible only once.  We observe that the performance of the merged model increases significantly as the amount of available data grows. Even when only 1\% of the data is visible once, using {\methodname} improves performance from 69.1 to 78.9, compared to the standard Task Arithmetic~\cite{TaskArithmetic_ICLR2023} method.

\begin{table*}[t]
\centering
\caption{Multi-task performance when merging eight ViT-B/32 models under the wild data.}
\vspace{-5pt}
\label{tab:performance_vitbase32_wild} 
\resizebox{\linewidth}{!}{  
\begin{tabular}{l|c|cccccccc|cc}
\hline
{Method}  & {Wild Data}  &  {SUN397}  &  {Cars}  &  {RESISC45}  &  {EuroSAT}  &  {SVHN}  &  {GTSRB}  &  {MNIST}  &  {DTD}  & \textbf{Avg.} $\uparrow$   \\
\hline
{Traditional MTL}   & -  &  73.9  &  74.4  &  93.9  & 98.2    &  95.8  &  98.9   &  99.5   & 77.9 & 88.9  \\
Task Arithmetic~\citep{TaskArithmetic_ICLR2023}  & - &55.2 &54.9 &66.7 &78.9 &80.2 & 69.7 &97.3 &50.4 & 69.1 \\
\hline
\rowcolor{mygray}
Task Arithmetic w/ {\cmethodname} (Ours)   & CIFAR100~\citep{cifar100}  & 52.1  & 52.7 & 67.0 &  62.5 & 82.1 & 63.6 & 95.6 & 45.2 & 65.1 \\
\rowcolor{mygray}
Task Arithmetic w/ 
{\cmethodname} (Ours) & ImageNet~\citep{deng2009imagenet}  & 62.6  & 56.5 & 70.8 &  74.2 & 79.4 & 63.5 & 96.1 & 58.2 &70.2 \\
\rowcolor{mygray}
\textbf{Task Arithmetic w/ \methodname} (Ours) & CIFAR100~\citep{cifar100}  &  65.4 & 50.2 & 77.7 & 87.5 & 95.0 & 93.5 &98.9 & 53.1 & 77.7 \\
\rowcolor{mygray}
\textbf{Task Arithmetic w/ \methodname} (Ours) & ImageNet~\citep{deng2009imagenet} & 72.2  & 63.6 &87.3  & 97.8 & 93.5 & 91.3 & 98.1 & 69.5  & 84.2 \\
\hline
\end{tabular}
}
\end{table*}

\begin{table*}[t]
\centering
\caption{Multi-task performance when performing {\cmethodname} at {different blocks}.}
\vspace{-5pt}
\label{tab:block_surgery_appendix} 
\resizebox{\linewidth}{!}{  
\begin{tabular}{l|c|cccccccc|c}
\hline
{Method}  & Block ID &  {SUN397}  &  {Cars}  &  {RESISC45}  &  {EuroSAT}  &  {SVHN}  &  {GTSRB}  &  {MNIST}  &  {DTD}  & \textbf{Avg.} $\uparrow$  \\
\hline
{Traditional MTL}  & -  &  73.9  &  74.4  &  93.9  & 98.2    &  95.8  &  98.9   &  99.5   & 77.9 & 88.9  \\
{Task Arithmetic}~\citep{TaskArithmetic_ICLR2023}  & -  &55.2 &54.9 &66.7 &78.9 &80.2 & 69.7 &97.3 &50.4 & 69.1 \\
\hline
\rowcolor{mygray}
{Task Arithmetic w/ {\cmethodname}} (Ours)    & Block \#1  &  56.4    &   57.1   &  68.9  & 88.0  & 84.6 & 71.6  & 97.6  & 51.2  & 71.9 \\
\rowcolor{mygray}
{Task Arithmetic w/ {\cmethodname}} (Ours)      &  Block \#3 &  48.7    &   50.6   &   69.8 &  88.1&   87.5 &67.4 &  97.6 & 50.5 &   70.0 \\
\rowcolor{mygray}
{Task Arithmetic w/ {\cmethodname}} (Ours)     & Block \#5  &  57.1    &  58.0    & 71.2   & 73.5 & 86.6 & 74.5  &  97.9 & 51.3 & 71.3 \\
\rowcolor{mygray}
{Task Arithmetic w/ {\cmethodname}} (Ours) &    Block \#7  &  57.2    &  53.5    & 78.0   & 96.4 & 83.1 & 75.6  & 98.8  & 54.9 &  74.7\\
\rowcolor{mygray}
{Task Arithmetic w/ {\cmethodname}} (Ours)   & Block \#9  &   58.0   &  44.6  &77.8  & 94.0 & 82.8  &  60.2 & 99.0 & 57.3  &  71.7  \\
\rowcolor{mygray}
{Task Arithmetic w/ {\cmethodname}} (Ours)     & Block \#11  &  57.7    &  54.2    &  76.6  & 92.9 &  81.8 & 70.4  & 98.0  & 58.7 &  73.8 \\
\rowcolor{mygray}
{Task Arithmetic w/ {\cmethodname}} (Ours)     & Final Representation  &   64.2   &   58.4   &   83.1 & 97.8  & 86.7 & 87.6  & 98.7  & 69.3 &80.7  \\
\hline
\rowcolor{mygray}
\textbf{Task Arithmetic w/ \methodname} (Ours)  
& All Blocks  & 73.8   & 67.9 & 94.5 & 99.6  & 96.8 & 98.8 & 99.5 & 78.0 & 88.6 \\
\hline
\end{tabular}
}
\end{table*}

\subsection{Surgery and SurgeryV2 in the Wild Data Setting}

The standard {\cmethodname} in $\S$\ref{subsec:surgery} and {\methodname} in $\S$\ref{subsec:ourmethod} require parameter optimization (i.e., Eq.~\ref{eq:surgery_v1} and Eq.~\ref{eq:surgery_v2} in $\S$\ref{sec:method}). However, unlabeled test data may not always be available during model merging. Therefore, we verified the feasibility of executing {\cmethodname} and {\methodname} on public datasets. As shown in Tab.~\ref{tab:performance_vitbase32_wild}, without loss of generality, we use two commonly used benchmark datasets in the vision domain, CIFAR100~\citep{cifar100} and ImageNet~\citep{deng2009imagenet}, as auxiliary data to perform the representation surgery (e.g., our {\cmethodname} or our {\methodname}).

The results show that, compared to Tab.~\ref{tab:performance_vitbase32} (in $\S$\ref{subsec:performance}) when using unlabeled test data, the performance of the {\cmethodname} or {\methodname} using unlabeled test data has decreased under wild data conditions. However, compared to traditional Task Arithmetic~\cite{TaskArithmetic_ICLR2023}, which does not employ representation surgery and achieves only an average accuracy of 69.1\%, the proposed {\methodname} attains an accuracy of 84.2\% on the ImageNet dataset. Under the CIFAR100 dataset, the performance will be lower, which may be due to the lower sample diversity and image resolution of CIFAR100 compared to ImageNet. Overall, our {\methodname} also achieved acceptable performance under wild data, greatly expanding the available scenarios of our {\methodname} solution.

\subsection{Component Analysis}
\label{subsec:analysis_appendix}

\textbf{Surgery in Different Blocks}.
The analysis in $\S$\ref{subsec:revisiting} shows that the `representation bias' phenomenon exists at each block of the merged model, prompting the question: {What is the performance when conducting representation surgery on individual blocks}?
For instance, ViT-B/32 has a total of 12 blocks, and we tested the performance of performing {\cmethodname} operations on the $\{1,3,5,7,9,11\}$-th block, the final layer (i.e., the vanilla {\cmethodname} scheme in $\S$\ref{subsec:surgery}), and all blocks (i.e., the {\methodname} scheme proposed in $\S$\ref{subsec:ourmethod}). 

As shown in Tab.~\ref{tab:block_surgery_appendix}, the general trend is that the benefits are more obvious when the {\cmethodname} is performed on the posterior block. For example, the average accuracy of performing {\cmethodname} on the 1st block is 71.9, while the accuracy of performing {\cmethodname} on the 11-th block is 73.8, and the accuracy of performing {\cmethodname} on the last block is $80.7$. Furthermore, our {\methodname} operates on all blocks, approaching performance (i.e., 88.6) closer to that of traditional MTL (i.e., 88.9).

\textbf{{\cmethodname} and {\methodname} with Different Loss function}.
The objective of the {\cmethodname} module and the {\methodname} module is to minimize the gap between the representation of the merged model and the representation of the individual expert model (Eq.~\ref{eq:surgery_v1} in $\S$\ref{subsec:surgery} and Eq.~\ref{eq:surgery_v2} in $\S$\ref{subsec:ourmethod}), thus making any distance-measuring function applicable. As shown in Tab.~\ref{tab:lossfun_vitbase32_appendix} below, we tested the performance of using $L_1$ loss, mean-square error (MSE), and negative cosine similarity (Neg. Cos.) as loss functions. 
We can observe that under different loss functions, our scheme is significantly better than Task Arithmetic, which does not apply the surgery scheme. In particular, our {\methodname} achieves performance close to that of traditional MTL, indicating that our scheme is robust to different loss functions.

\begin{table*}[t]
\centering
\caption{Multi-task performance on the ViT-B/32 model with {different loss functions}.
}
\label{tab:lossfun_vitbase32_appendix} 
\vspace{-5pt}
\resizebox{\linewidth}{!}{  
\begin{tabular}{l|c|cccccccc|cc}
\hline
{Method}  & Loss Function  &  {SUN397}  &  {Cars}  &  {RESISC45}  &  {EuroSAT}  &  {SVHN}  &  {GTSRB}  &  {MNIST}  &  {DTD}  & \textbf{Avg.} $\uparrow$  \\
\hline
{Traditional MTL}  & -  &  73.9  &  74.4  &  93.9  & 98.2    &  95.8  &  98.9   &  99.5   & 77.9 & 88.9  \\
Task Arithmetic~\citep{TaskArithmetic_ICLR2023} &-   &55.2 &54.9 &66.7 &78.9 &80.2 & 69.7 &97.3 &50.4 & 69.1 \\
\hline
\rowcolor{mygray}
{Task Arithmetic w/ {\cmethodname}} (Ours) & L1 Loss  &63.8 &59.9 &83.3 &97.9 &87.0 &87.0 &98.6 &69.4 &80.9\\
\rowcolor{mygray}
{Task Arithmetic w/ {\cmethodname}} (Ours) &MSELoss & 64.3 & 59.8 & 84.0 &97.8  &87.6  & 88.7 & 98.8   & 69.8 & 81.4 \\
\rowcolor{mygray}
{Task Arithmetic w/ {\cmethodname}} (Ours) & Neg. Cos. & 63.8  & 59.2 & 84.5  & 97.5 & 88.8 &  86.4 & 99.0 &  70.3 & 81.2 \\
\hline
\rowcolor{mygray}
\textbf{Task Arithmetic w/ \methodname} (Ours) & L1 Loss  & 73.8   & 67.9 & 94.5 & 99.6  & 96.8 & 98.8 & 99.5 & 78.0 & 88.6 \\
\rowcolor{mygray}
\textbf{Task Arithmetic w/ \methodname} (Ours) & MSE Loss & 74.0 & 70.2 & 94.7  & 99.4 & 95.9  & 98.3  & 99.4 &  77.6  & 88.7  \\
\rowcolor{mygray}
\textbf{Task Arithmetic w/ \methodname} (Ours) & Neg. Cos. &74.3  & 70.6 & 94.7  & 99.8 & 96.3  & 98.6  &99.6  &  77.8  & 88.9 \\
\hline
\end{tabular}
}
\end{table*}


\begin{table*}[t]
\centering
\caption{Impact of the amount of {rank} on model performance in {\cmethodname} and {\methodname} modules.}
\vspace{-5pt}
\label{tab:surgeryv2_rank_appendix} 
\resizebox{\linewidth}{!}{  
\begin{tabular}{l|c|cccccccc|c}
\hline
{Method}  & Rank ($r$) &  {SUN397}  &  {Cars}  &  {RESISC45}  &  {EuroSAT}  &  {SVHN}  &  {GTSRB}  &  {MNIST}  &  {DTD}  & \textbf{Avg.} $\uparrow$  \\
\hline
{Traditional MTL}  & -  &  73.9  &  74.4  &  93.9  & 98.2    &  95.8  &  98.9   &  99.5   & 77.9 & 88.9  \\
{Task Arithmetic}~\citep{TaskArithmetic_ICLR2023}  & - &55.2 &54.9 &66.7 &78.9 &80.2 & 69.7 &97.3 &50.4 & 69.1 \\
\hline
\rowcolor{mygray}
{Task Arithmetic w/ {\cmethodname}} (Ours)& $4$ &62.6 &55.9 &76.7 &78.7 &83.4 &79.6 &97.7 &60.0 &74.3\\
\rowcolor{mygray}
{Task Arithmetic w/ {\cmethodname}} (Ours)& $8$ &63.2 &58.5 &79.9 &93.9 &84.5 &82.1 &98.5 &64.2 &78.1\\
\rowcolor{mygray}
{Task Arithmetic w/ {\cmethodname}} (Ours)& $16$  &63.8 &59.9 &83.3 &97.9 &87.0 &87.0 &98.6 &69.4 &80.9\\
\rowcolor{mygray}
{Task Arithmetic w/ {\cmethodname}} (Ours)& $32$ &64.7 &62.4 & 87.3 &98.3 &88.3 &92.8 &98.8 &72.5 &83.1\\
\rowcolor{mygray}
{Task Arithmetic w/ {\cmethodname}} (Ours)& $64$ &65.6 &62.7 &89.8 &98.3 & 88.7 &95.6 &98.9 &74.7 &84.3\\
\hline
\rowcolor{mygray}
\textbf{Task Arithmetic w/ \methodname} (Ours) & $4$ &  72.7  &   68.1    &  93.8   &  99.5   &   96.4 &  98.1  & 99.6  & 76.3 & 88.1 \\
\rowcolor{mygray}
\textbf{Task Arithmetic w/ \methodname} (Ours) & $8$  & 73.6   &   67.6    &  93.7   &  99.7   &  96.6  &  98.4  &  99.5 & 78.2 &88.4  \\
\rowcolor{mygray}
\textbf{Task Arithmetic w/ \methodname} (Ours) & $16$ & 73.8   & 67.9 & 94.5 & 99.6  & 96.8 & 98.8 & 99.5 & 78.0 & 88.6 \\
\rowcolor{mygray}
\textbf{Task Arithmetic w/ \methodname} (Ours) & $32$ & 74.5   &  67.4     &  94.6   &  99.6   &  96.4  & 98.8   & 99.5  & 77.7 &  88.6 \\
\rowcolor{mygray}
\textbf{Task Arithmetic w/ \methodname} (Ours) & $64$  &  74.3  &  72.1     & 94.4    &   99.7  &  96.9  & 98.2   & 99.4  & 77.6 & 89.1 \\
\hline
\end{tabular}
}
\end{table*}

\begin{figure*}[t]
    \centering 
     \includegraphics[width=0.24\textwidth]{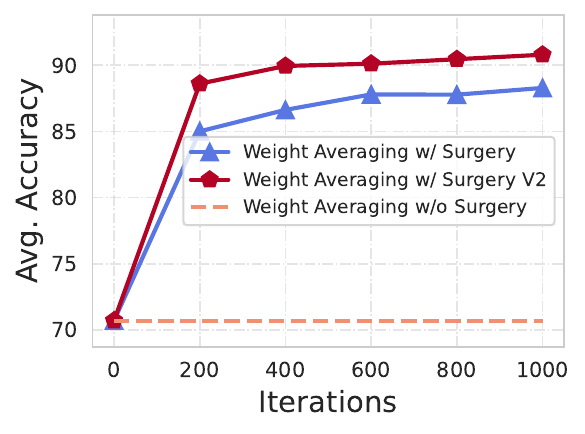}
     \includegraphics[width=0.24\textwidth]{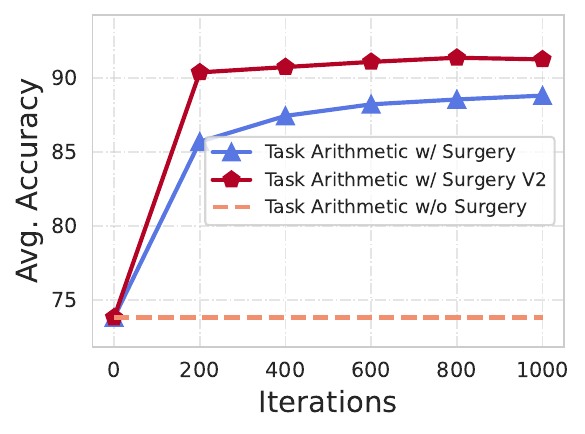}
      \includegraphics[width=0.24\textwidth]{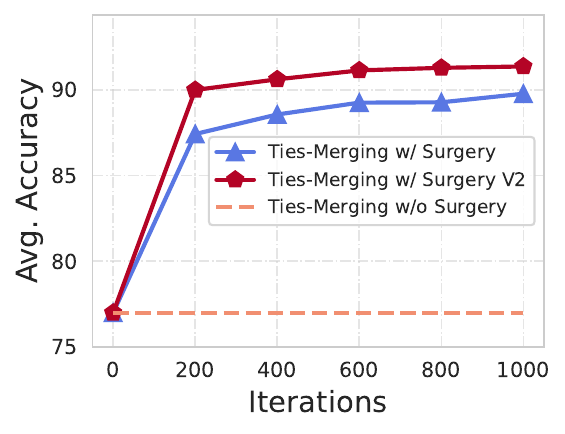}
       \includegraphics[width=0.24\textwidth]{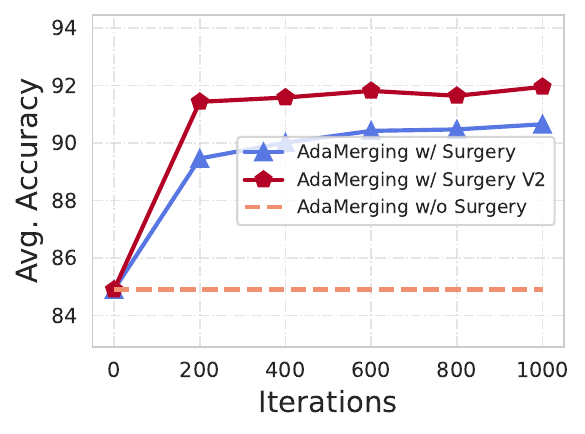}
    \vspace{-5pt}
    \caption{An illustration of how the performance of the merged model changes w.r.t the number of iterations when merging eight \textbf{ViT-B/16} models. From left to right are model merging solutions based on Weight Averaging, Task Arithmetic, Ties-Merging, and AdaMerging.}  
\label{fig:iter_acc_vit-b16_appendix} 
\vspace{-10pt}
\end{figure*}

\textbf{{\cmethodname} and {\methodname} with Different Ranks}. 
One hyperparameter in {\cmethodname} and {\methodname} is rank (i.e., $r$ of Eq.~\ref{eq:surgery} in $\S$\ref{subsec:surgery} or Eq.~\ref{eq:surgery_module_v2} in $\S$\ref{subsec:ourmethod}), which reflects the capacity of the {\cmethodname} module. As shown in Tab.~\ref{tab:surgeryv2_rank_appendix} below, we evaluated the performance changes for various values of the rank $r$ (e.g., $\small r \in \{4,6,8,16,32,64\}$). We observe that the performance of the merged model improves as the rank increases. In particular, {\methodname} is very close to the performance of the traditional MTL model at different ranks.

\textbf{Performance w.r.t. Iterations}.
Illustrated in Fig.~\ref{fig:iter_acc_vit-b16_appendix}, we show how the accuracy of our {\cmethodname} and our {\methodname} changes during iterations when merging eight ViT-B/16 models. It is evident that the {\methodname} method attains superior performance with fewer iterations compared to the vanilla {\cmethodname} method, ultimately yielding better final performance.

\subsection{Visualization}
\label{subsec:Visualization}

\begin{figure*}[t]
    \centering 
     \includegraphics[width=0.245\textwidth]{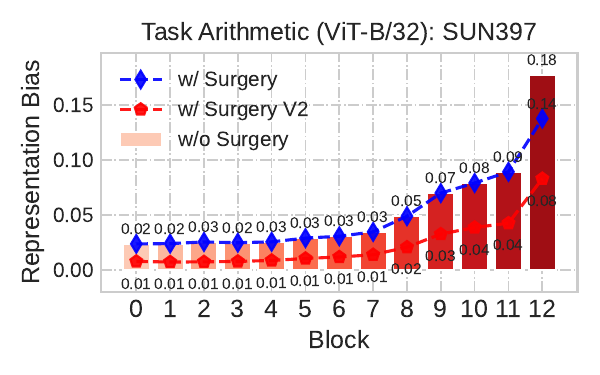}
     \includegraphics[width=0.245\textwidth]{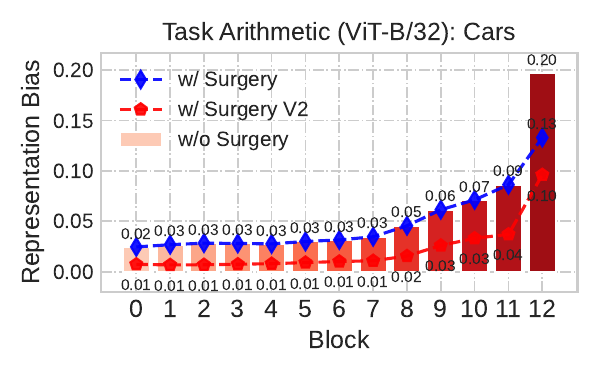}
     \includegraphics[width=0.245\textwidth]{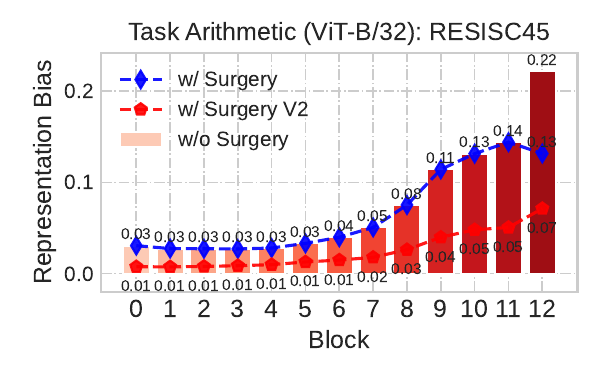}
     \includegraphics[width=0.245\textwidth]{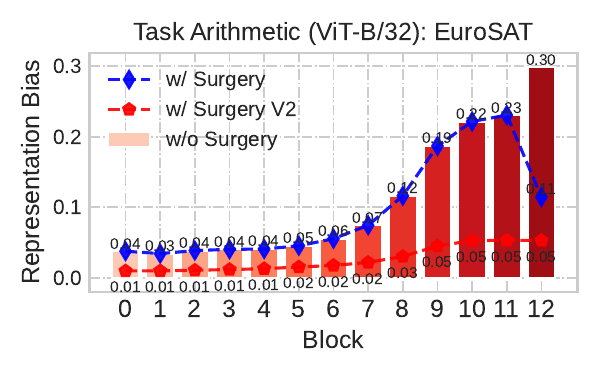}
      \includegraphics[width=0.245\textwidth]{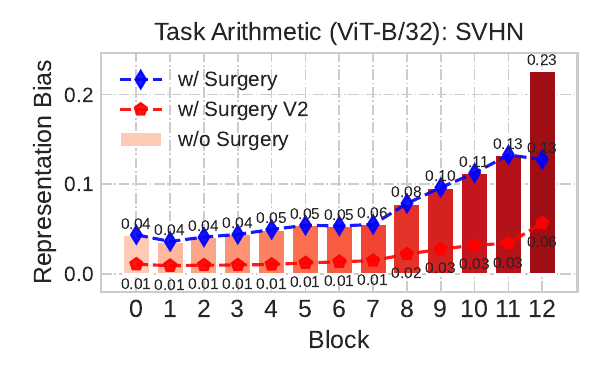}
     \includegraphics[width=0.245\textwidth]{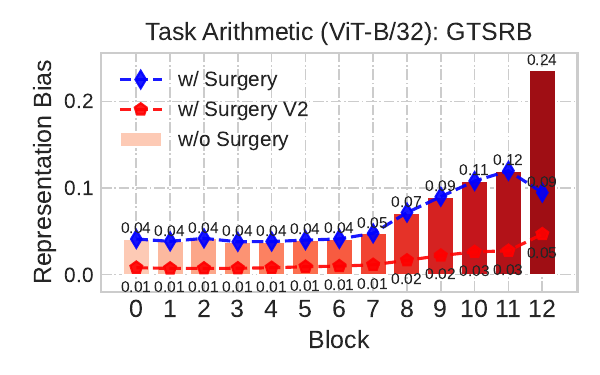}
     \includegraphics[width=0.245\textwidth]{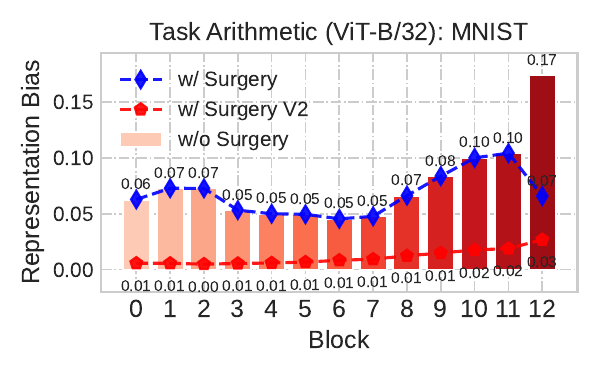}
     \includegraphics[width=0.245\textwidth]{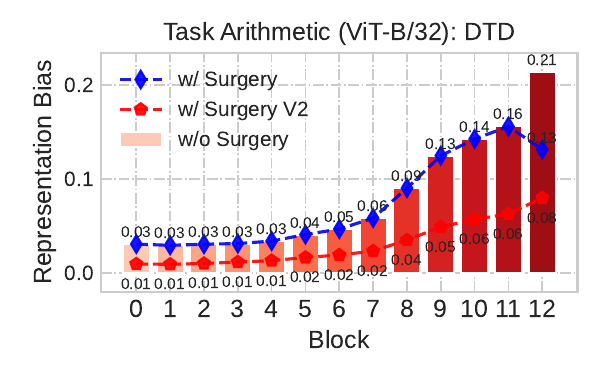}
    \caption{
    Visualization of the distribution of extracted representations (final layer) for individual models (\textcolor{blue}{blue}) and merged models (\textcolor{red}{red}) on eight datasets.
    The \textit{first column} represents the merged model obtained by \textbf{Task Arithmetic}~\cite{TaskArithmetic_ICLR2023} (ViT-B/32) without using the representation surgery scheme. The \textit{second column} indicates the use of the {\cmethodname} scheme in $\S$\ref{subsec:surgery}, and the \textit{third column} indicates the use of the {\methodname} scheme proposed in $\S$\ref{subsec:ourmethod}.
    }  
\label{fig:compare_surgery_v1v2_layerwise_bias_tv_vit_b32_appendix} 
\end{figure*}

\begin{figure*}[t]
    \centering 
     \includegraphics[width=0.245\textwidth]{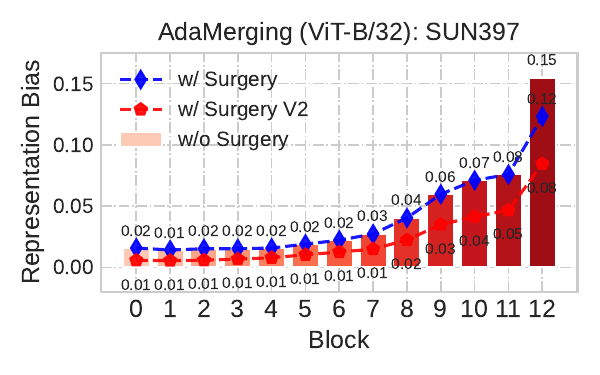}
     \includegraphics[width=0.245\textwidth]{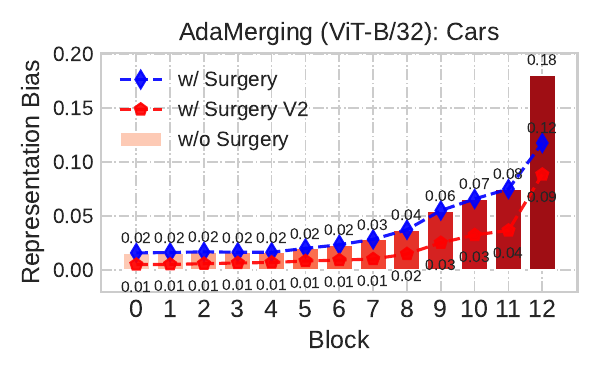}
     \includegraphics[width=0.245\textwidth]{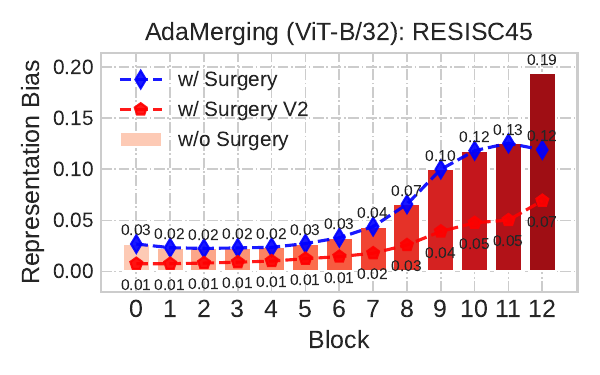}
     \includegraphics[width=0.245\textwidth]{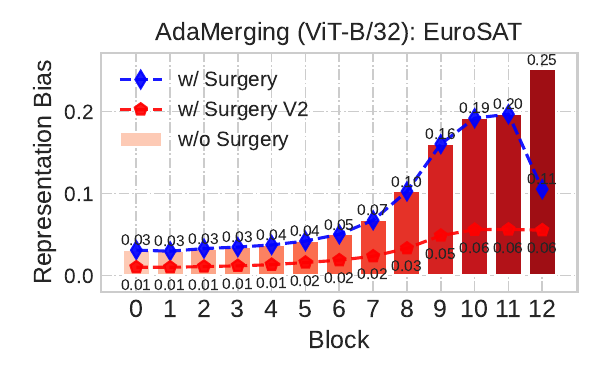}
      \includegraphics[width=0.245\textwidth]{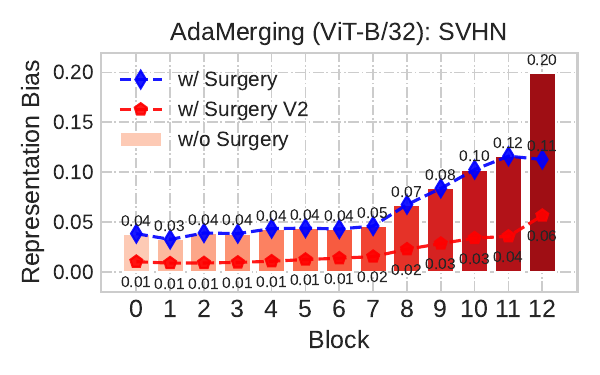}
     \includegraphics[width=0.245\textwidth]{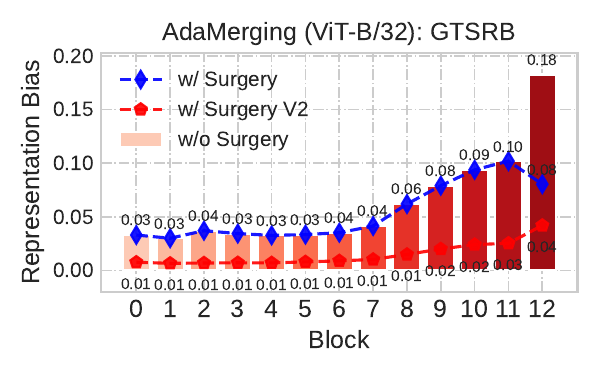}
     \includegraphics[width=0.245\textwidth]{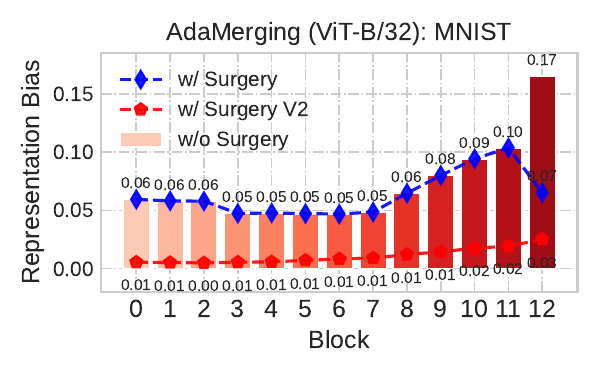}
     \includegraphics[width=0.245\textwidth]{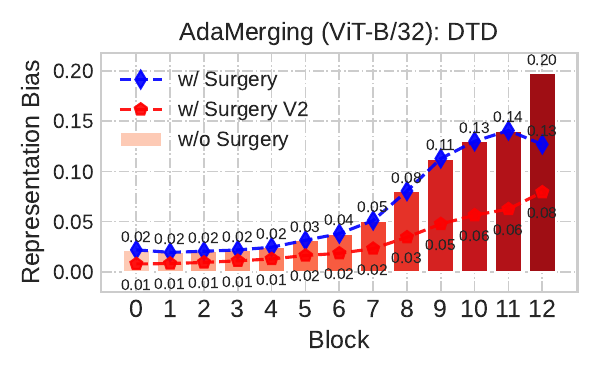}
    \vspace{-10pt}
    \caption{
    Visualization of the distribution of extracted representations (final layer) for individual models (\textcolor{blue}{blue}) and merged models (\textcolor{red}{red}) on eight datasets.
    The \textit{first column} represents the merged model obtained by \textbf{AdaMerging}~\cite{AdaMerging_ICLR2024} (ViT-B/32) without using the representation surgery scheme. The \textit{second column} indicates the use of the {\cmethodname} scheme in $\S$\ref{subsec:surgery}, and the \textit{third column} indicates the use of the {\methodname} scheme proposed in $\S$\ref{subsec:ourmethod}.
    }  
\label{fig:compare_surgery_v1v2_layerwise_bias_adamerging_vit_b32_appendix} 
\end{figure*}

\textbf{Representation Bias}.
In Fig.~\ref{fig:compare_surgery_v1v2_layerwise_bias_tv_vit_b32_appendix} and Fig.~\ref{fig:compare_surgery_v1v2_layerwise_bias_adamerging_vit_b32_appendix}, we show the block-wise representation bias on eight datasets, comparing two model merging techniques (i.e., Task Arithmetic~\cite{TaskArithmetic_ICLR2023} and AdaMerging~\cite{AdaMerging_ICLR2024}) under three conditions: without any surgery, with the original {\cmethodname} as proposed in $\S$\ref{subsec:surgery}, and with our {\methodname} in $\S$\ref{subsec:ourmethod}.

\begin{figure*}[t]
    \centering 
     \includegraphics[width=0.32\textwidth]{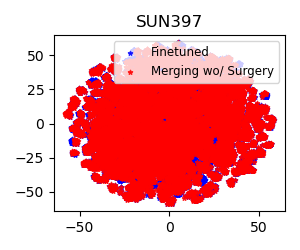}
     \includegraphics[width=0.32\textwidth]{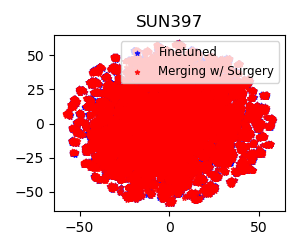}
     \includegraphics[width=0.32\textwidth]{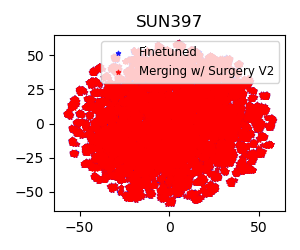}
      \includegraphics[width=0.32\textwidth]{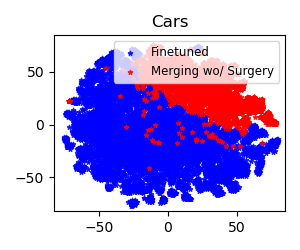}
     \includegraphics[width=0.32\textwidth]{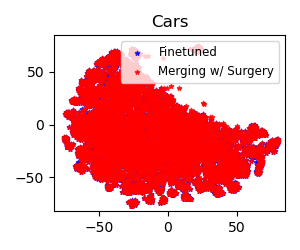}
     \includegraphics[width=0.32\textwidth]{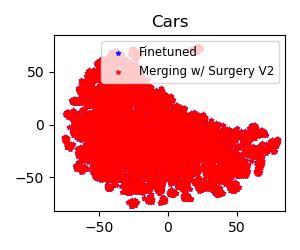}
      \includegraphics[width=0.32\textwidth]{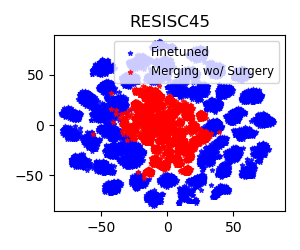}
     \includegraphics[width=0.32\textwidth]{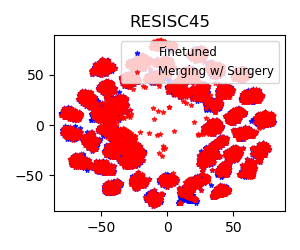}
     \includegraphics[width=0.32\textwidth]{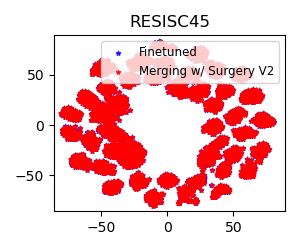}
      \includegraphics[width=0.32\textwidth]{images/distribution/task_arithmetic_ViT-B-32_EuroSAT_finetuned_Merging_wo_Surgery.png}
     \includegraphics[width=0.32\textwidth]{images/distribution/task_arithmetic_ViT-B-32_EuroSAT_finetuned_Merging_w_Surgery_V1.png}
     \includegraphics[width=0.32\textwidth]{images/distribution/task_arithmetic_ViT-B-32_EuroSAT_finetuned_Merging_w_Surgery_V2.png}
    \caption{Visualization of the distribution of extracted representations (final layer) for individual models (\textcolor{blue}{blue}) and merged model (\textcolor{red}{red}) in the setup of \textbf{ViT-B/32 architecture and Task Arithmetic method}.
    The \textit{first column} represents the merged model obtained by Task Arithmetic~\cite{TaskArithmetic_ICLR2023} without using the representation surgery scheme. The \textit{second column} indicates the use of the {\cmethodname} scheme in $\S$\ref{subsec:surgery}, and the \textit{third column} indicates the use of the {\methodname} scheme proposed in $\S$\ref{subsec:ourmethod}.}  
\label{fig:distribution_tv_vit_b32_part1_appendix} 
\end{figure*}

\begin{figure*}[t]
    \centering 
      \includegraphics[width=0.32\textwidth]{images/distribution/task_arithmetic_ViT-B-32_SVHN_finetuned_Merging_wo_Surgery.png}
     \includegraphics[width=0.32\textwidth]{images/distribution/task_arithmetic_ViT-B-32_SVHN_finetuned_Merging_w_Surgery_V1.png}
     \includegraphics[width=0.32\textwidth]{images/distribution/task_arithmetic_ViT-B-32_SVHN_finetuned_Merging_w_Surgery_V2.png}
      \includegraphics[width=0.32\textwidth]{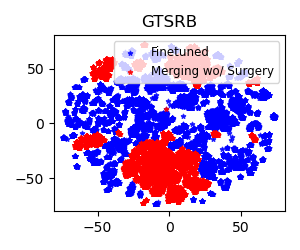}
     \includegraphics[width=0.32\textwidth]{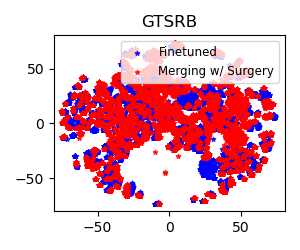}
     \includegraphics[width=0.32\textwidth]{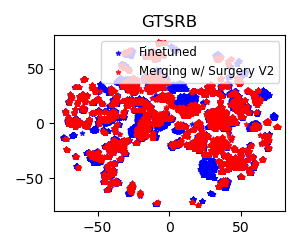}
      \includegraphics[width=0.32\textwidth]{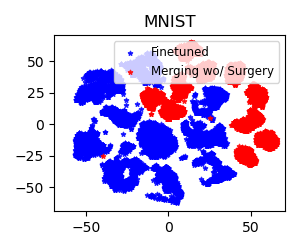}
     \includegraphics[width=0.32\textwidth]{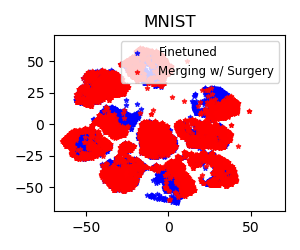}
     \includegraphics[width=0.32\textwidth]{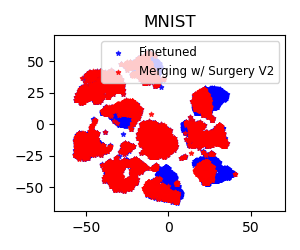}
      \includegraphics[width=0.32\textwidth]{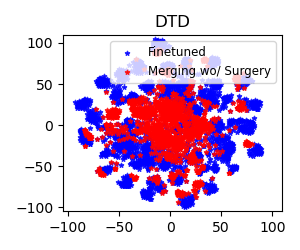}
     \includegraphics[width=0.32\textwidth]{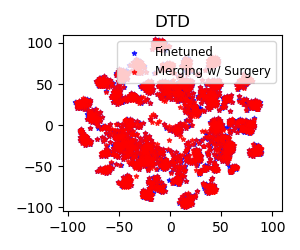}
     \includegraphics[width=0.32\textwidth]{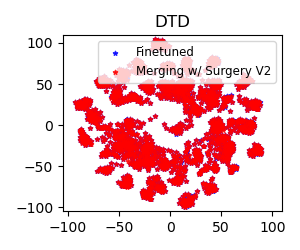}
    \caption{Visualization of the distribution of extracted representations (final layer) for individual models (\textcolor{blue}{blue}) and merged model (\textcolor{red}{red})  in the setup of \textbf{ViT-B/32 architecture and Task Arithmetic method}.
    The \textit{first column} represents the merged model obtained by Task Arithmetic~\cite{TaskArithmetic_ICLR2023} without using the representation surgery scheme. The \textit{second column} indicates the use of the {\cmethodname} scheme in $\S$\ref{subsec:surgery}, and the \textit{third column} indicates the use of the {\methodname} scheme proposed in $\S$\ref{subsec:ourmethod}.}  
\label{fig:distribution_tv_vit_b32_part2_appendix} 
\end{figure*}

\begin{figure*}[t]
    \centering 
     \includegraphics[width=0.32\textwidth]{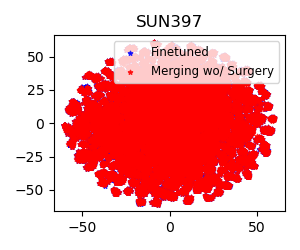}
     \includegraphics[width=0.32\textwidth]{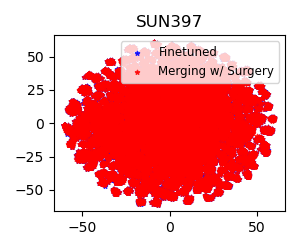}
     \includegraphics[width=0.32\textwidth]{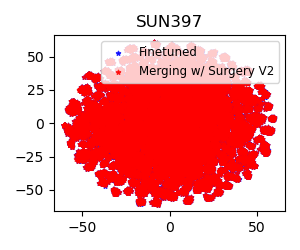}
      \includegraphics[width=0.32\textwidth]{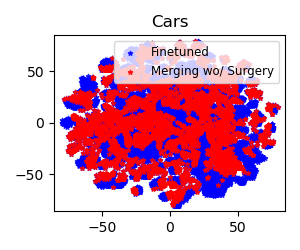}
     \includegraphics[width=0.32\textwidth]{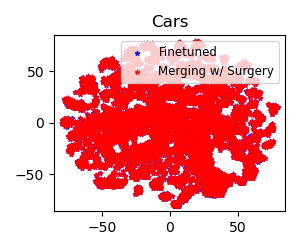}
     \includegraphics[width=0.32\textwidth]{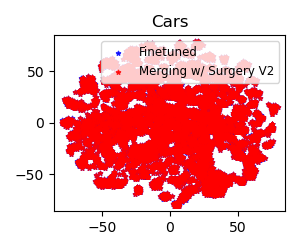}
      \includegraphics[width=0.32\textwidth]{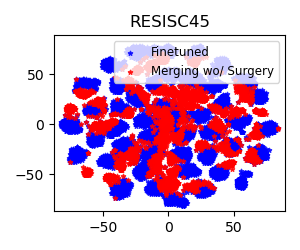}
     \includegraphics[width=0.32\textwidth]{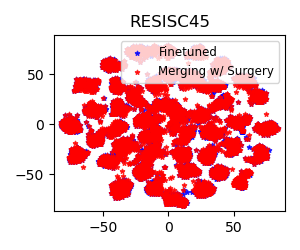}
     \includegraphics[width=0.32\textwidth]{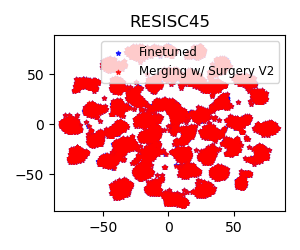}
      \includegraphics[width=0.32\textwidth]{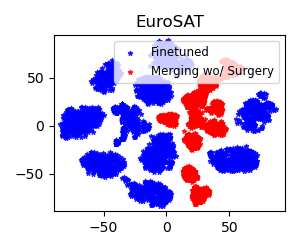}
     \includegraphics[width=0.32\textwidth]{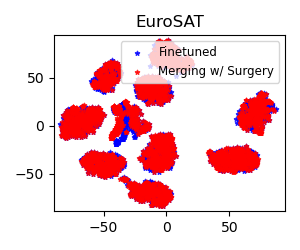}
     \includegraphics[width=0.32\textwidth]{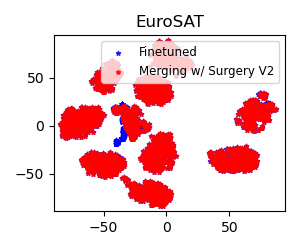}
    \caption{Visualization of the distribution of extracted representations (final layer) for individual models (\textcolor{blue}{blue}) and merged model (\textcolor{red}{red}) in the setup of \textbf{ViT-B/32 architecture and AdaMerging method}.
    The \textit{first column} represents the merged model obtained by AdaMerging~\cite{AdaMerging_ICLR2024} without using the representation surgery scheme. The \textit{second column} indicates the use of the {\cmethodname} scheme in $\S$\ref{subsec:surgery}, and the \textit{third column} indicates the use of the {\methodname} scheme proposed in $\S$\ref{subsec:ourmethod}.}   
\label{fig:distribution_adamerging_vit_b32_part1_appendix} 
\end{figure*}

\begin{figure*}[t]
    \centering 
      \includegraphics[width=0.32\textwidth]{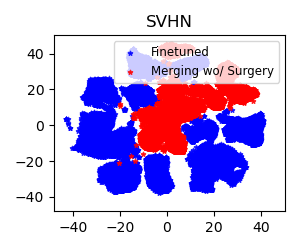}
     \includegraphics[width=0.32\textwidth]{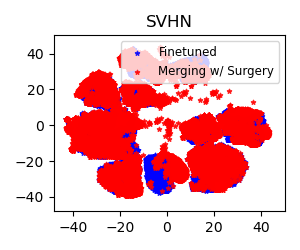}
     \includegraphics[width=0.32\textwidth]{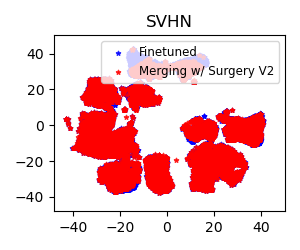}
      \includegraphics[width=0.32\textwidth]{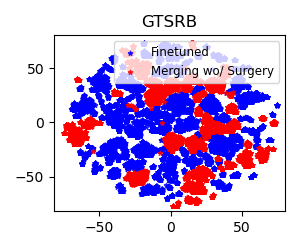}
     \includegraphics[width=0.32\textwidth]{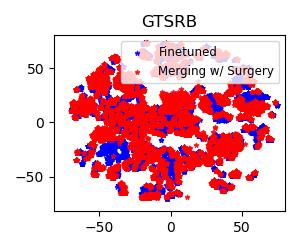}
     \includegraphics[width=0.32\textwidth]{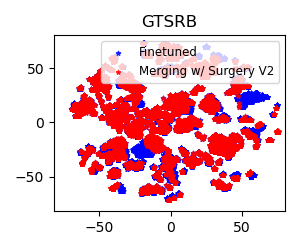}
      \includegraphics[width=0.32\textwidth]{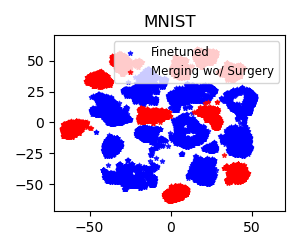}
     \includegraphics[width=0.32\textwidth]{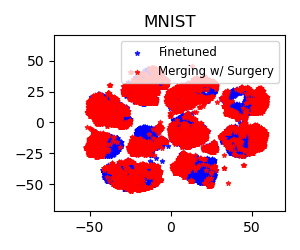}
     \includegraphics[width=0.32\textwidth]{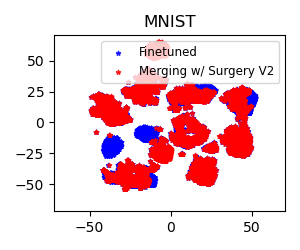}
      \includegraphics[width=0.32\textwidth]{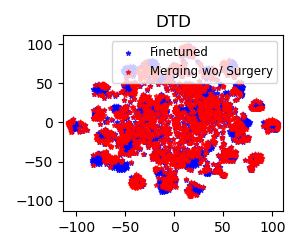}
     \includegraphics[width=0.32\textwidth]{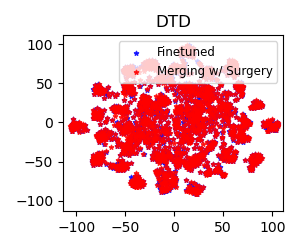}
     \includegraphics[width=0.32\textwidth]{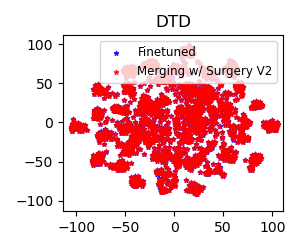}
    \caption{Visualization of the distribution of extracted representations (final layer) for individual models (\textcolor{blue}{blue}) and merged model (\textcolor{red}{red}) in the setup of \textbf{ViT-B/32 architecture and AdaMerging method}.
    The \textit{first column} represents the merged model obtained by AdaMerging~\cite{AdaMerging_ICLR2024} without using the representation surgery scheme. The \textit{second column} indicates the use of the {\cmethodname} scheme in $\S$\ref{subsec:surgery}, and the \textit{third column} indicates the use of the {\methodname} scheme proposed in $\S$\ref{subsec:ourmethod}.} 
\label{fig:distribution_adamerging_vit_b32_part2_appendix} 
\end{figure*}

\textbf{Representation Distribution}.
As shown in Fig.~\ref{fig:distribution_tv_vit_b32_part1_appendix} and Fig.~\ref{fig:distribution_tv_vit_b32_part2_appendix}, we visualize the representation distributions of the original Task Arithmetic~\cite{TaskArithmetic_ICLR2023}, the merged model that executes our {\cmethodname} or our {\methodname} on eight datasets. Likewise, Fig.~\ref{fig:distribution_adamerging_vit_b32_part1_appendix} and Fig.~\ref{fig:distribution_adamerging_vit_b32_part2_appendix} represent visualizations of the AdaMerging~\cite{AdaMerging_ICLR2024} method.

\end{document}